\documentclass[
journal=nlp,
manuscript=research-article,
year=2026,
]{cup-journal}

\usepackage{amsmath}
\usepackage[nopatch]{microtype}
\usepackage{booktabs}

\usepackage{xurl}

\usepackage{graphicx}

\usepackage{subcaption}

\usepackage{multirow}

\usepackage{fontspec}
\newfontface\engbold{texgyrepagella-bold.otf}[Path=./]
\newfontface\engital{texgyrepagella-italic.otf}[Path=./]
\newfontface\engbi{texgyrepagella-bolditalic.otf}[Path=./]

\newcommand{\bfeng}[1]{{\engbold #1}}
\newcommand{\iteng}[1]{{\engital #1}}

\newfontfamily\devanagarifont[Script=Devanagari]{Lohit_Nepali_Regular.ttf}

\newcommand{\nep}[1]{{\devanagarifont #1}}

\title{From Promises to Accountability: Comparative AI Modeling and Visualization of Electoral Commitments in Nepali Political Texts}

\author{Prakash Poudyal}
\affiliation{Information and Language Processing Research Lab, Kathmandu University, Kavre, Nepal}
\email{prakash@ku.edu.np}

\author{Nischal Karki}
\affiliation{Information and Language Processing Research Lab, Kathmandu University, Kavre, Nepal}

\alsoaffiliation{Joint first authors}

\author{Amrit Dahal}
\affiliation{Department of Artificial Intelligence, Kathmandu University, Nepal}

\author{Rajani Chulyadyo}
\affiliation{Department of Computer Science and Engineering, Kathmandu University, Nepal}

\author{Sameer Tamrakar}
\affiliation{Department of Computer Science and Engineering, Kathmandu University, Nepal}

\author{Bal Krishna Bal}
\affiliation{Information and Language Processing Research Lab, Kathmandu University, Kavre, Nepal}


\begin{document}
\begin{abstract}
Election Commitment Mining refers to the analysis of election commitments made by electoral candidates or political parties during election campaigns, directed toward voters and covering various development agendas. Candidates are expected to fulfill these commitments once they are elected and assume office. However, many elected candidates and political parties, especially in developing South Asian countries, do not fully deliver on the promises they make before elections. Even after the elections, commitments presented through election manifestos, posters, pamphlets, videos, and other campaign materials are often not systematically or transparently archived. As a result, these promises tend to fade away over time, making it difficult to hold the elected candidates accountable for their commitments. To address this persistent issue, this research aims to develop a system capable of structuring the commitments made by candidates during election periods. In this work, we created the first gold-standard dataset of Nepali political texts (Cohen’s $\kappa$ = 0.777) and developed as well as evaluated models through a comparative analysis of machine learning, deep learning and large language models to automatically extract and analyze election commitments, ultimately determining the best-performing model for this task. Our evaluation showed that LaBSE-LR achieved the best results among machine learning models (0.897 accuracy, 0.896 macro $F_1$-score, 0.947 AUROC), while fastText-LSTM achieved the highest performance among deep learning models (0.914 accuracy, 0.913 macro $F_1$-score, 0.961 AUROC). Among the three evaluated large language models, Qwen3-8B performed best in the few-shot setting (0.848 macro $F_1$-score). The outcome of this work is integrated into a mobile application to ensure accessibility for the general public. The commitments made by candidates can be visualized in the mobile app with the support of GAI, further enhancing transparency, civic engagement, and electoral accountability.
\end{abstract}

\section{Introduction}

\label{sec:introduction}

During election campaigns, candidates make commitments in various areas such as education, health, economy, and other development agendas. These electoral commitments are communicated through mass meetings, public speeches, election manifestos, pamphlets, television and radio interviews, social media, and other platforms. After the election, candidates who receive the majority of votes are elected as the political representatives of their respective constituencies. If an elected candidate is highly influential and represents a major political party, they may play a significant role in forming the government. Candidates are expected to fulfill the commitments made during the election campaign once they assume office. However, many elected candidates and parties in developing South Asian countries such as Nepal \autocite{pageni2014political, paudel2020local} and Pakistan \autocite{farrukh2021portrayal} pursue power and position rather than fulfilling their promises. These commitments in their manifestos are often not reflected in budget planning or government policies. Neither the opposition candidates who lost the election nor the citizens typically question the promises and commitments made by elected representatives before the election. One major reason for this is that citizens are often not in a position to evaluate whether the commitments made by candidates have been fulfilled. In Nepal, once the election concludes, candidates’ commitments, expressed through manifestos, posters, pamphlets, videos, and other campaign materials, are rarely preserved or systematically archived. As a result, these promises gradually fade from public discourse. The absence of organized documentation and monitoring mechanisms makes it challenging for citizens to evaluate the fulfillment of these commitments and highlights the need for more structured accountability systems.

In recent years, due to advances in Natural Language Processing (NLP), political texts, manifestos, and public discourse can be automatically analyzed at scale revealing the representatives' commitments, ideologies, and policy agendas. There have been multiple studies that have applied such methods to analyze commitments, semantics, and topics in political texts globally \citep{DBLP:conf/icwsm/OConnorBRS10, hirst2010partystatus, thomas2012votedeterminingsupportopposition, gottipati-etal-2013-learning, article, 46934237-f0af-3c79-8cb9-4fcc1e16944e, Grimmer_2010, sevenans2014automated, menini-etal-2017-topic, glavas-etal-2016-unsupervised, Hillard15052008, Collingwood01072012, karan-etal-2016-analysis, glavas-etal-2017-cross, glavas-etal-2017-unsupervised, subramanian-etal-2018-hierarchical, 10.1007/978-3-031-74491-4_8, Fagni_2022}. However, such approaches have not yet been applied in Nepal, especially for low-resource languages like Nepali. 

Nepali is an Indo-Aryan language of the Indo-Iranian branch of the Indo-European family of languages. It is spoken mainly in Nepal and also in parts of India and of Bhutan. Nepali is written in the Devanagari script which it has in common with languages like Hindi, Marathi and Sanskrit. It generally has subject-object-verb (SOV) word order and uses postpositions rather than prepositions. The language also has features such as rich verb inflection, case marking, honorific distinctions, and productive verb compounds \autocite{bal2004structure}. Nepali is closely related to other Indo-Aryan languages, particularly Hindi and other languages of northern South Asia, though it has its own grammatical, lexical and phonological features. Nepali is an official language spoken by a large number of people in Nepal, but in natural language processing it is still relatively low-resource \autocite{shahi2022natural, rajan2021survey}. In comparison with English and other high resource languages, there are fewer large and diverse corpora for Nepali, manually annotated datasets, resources for parallel translation, evaluation benchmarks and pre-trained language models \autocite{shahi2022natural}. This resource limitation is particularly important for machine translation and speech-act classification, where the availability of high-quality labelled data strongly affects model training and evaluation \autocite{chaudhary2020efforts, lamichhane2023give}. Therefore, in this paper, we use ``low-resource'' to refer mainly to the limited availability of computational and annotated resources rather than to the number of Nepali speakers.

Leveraging the global studies and advances in NLP, we address the problem in the political context of Nepal by creating an annotated gold-standard dataset of electoral commitments and developing a model through comparative benchmarking of various Machine Learning (ML) models, Deep Learning (DL) models, and Large Language Models (LLMs) to automatically extract and analyze publicity materials released by election candidates. In addition, we leverage Generative AI (GAI) models to generate visual representations of electoral commitments, making abstract promises more tangible and engaging for the public. These generated images, along with commitment summaries, are then delivered through a mobile application to ensure accessibility to the general public and to promote civic engagement and electoral accountability.

The rest of the paper is organized as follows: the literature review is presented in Section \ref{sec:literature-reviews}, the methodology of our work is detailed in Section \ref{sec:methods and materials}, all experiments and their results along with the ablation studies and error analysis of the models are explained in Section \ref{sec:experiments_results}, the discussion of our findings is discussed in Section \ref{sec:discussion}, followed by the limitations and ethical considerations in Sections \ref{sec:limitations} and \ref{sec:ethics}, respectively. Finally, the conclusion and our future plans to advance this research are presented in Section \ref{sec:conclusion}.

\section{Literature Reviews}
\label{sec:literature-reviews}

The role of political communication in shaping electoral outcomes has been an important subject of academic investigation in different political settings. While numerous studies have analyzed political debates, there is limited research specifically examining commitments made by political actors. In the case of Nepal, \citet{lamichhane2023give} explore the role played by social media such as Facebook in election-related political communication and the extent to which politicians utilized Facebook as a direct platform to connect with voters. They report that Facebook is the most popular medium for political use in Nepal and it was one of the most widely used communication mediums during the 2022 local elections. They further analyze Facebook posts of mayoral candidates where they find a predominance of assertive speech acts and a limited use of declarative speech acts. They also find no fundamental difference between winning and losing candidates in their use of speech acts, which they interpret as demonstrating a habitual reporting style rather than the strategic deployment of persuasive logical or emotional appeals. This shows that the politicians do not understand the strategic power of speech acts in political discourse.

\citet{paudel2020local} investigates the electoral accountability of local representatives in two districts of Nepal, Sindhuli and Surkhet, on the basis of promises made before elections. The authors collected 455 questionnaires from citizens and conducted 28 interviews with chairpersons, vice-chairpersons, mayors, deputy mayors and executive officers of rural municipalities and municipalities using an open-ended questionnaire. They find that the performance of local governments in Surkhet has a stronger significant relationship with an electoral accountability index than in Sindhuli. However, overall citizens were neither clearly satisfied nor dissatisfied with local representatives. The study finds that electoral accountability in Nepal is still a work-in-progress, with representatives, employes and the public still learning to manage its processes.

Similarly, \citet{dahal2001electoral} states that the electoral system should offer chances for people to participate in political process effectively. He also observes that voters often exhibit bandwagon-type behavior, where they support candidates and parties that seem likely to win rather than assessing them on objective grounds. In order to resolve this, he argues that the civic fabric should be strengthened through education and support mechanisms that can raise political consciousness and help citizens make more informed electoral choices.

Beyond Nepal, \citet{dama2021analysis} studies the electoral pattern in Turkey since the introduction of the multi-party system in 1946. Despite disruptions caused by military coups in 1960 and 1980 and a failed coup attempt in 2016, parliamentary elections between 2002 and 2015 saw consistently high voter turnout, reflecting strong public commitment to democratic participation. During this period, the Justice and Development Party (AKP) dominated politics, strategically incorporating social policy commitments such as poverty alleviation, healthcare, education, and employment into its election manifestos. Content analysis of these manifestos highlights the role of social protection as a political tool for maintaining voter support in a shifting political and economic environment.

The role of political parties in making specific, localized commitments is highlighted by \citet{Ashworth_2000}, who examines how pledges are framed around local issues and how accountability mechanisms function in fulfilling them. Manifestos are presented not only as declarations of intent but also as tools enabling voters to hold parties accountable. This perspective aligns with the broader literature on electoral mandates such as \citep{Budge_Hofferbert_1990}, which explores how party platforms translate into governance and public expenditure.  \citet{Ashworth_2000} focus on local elections further emphasizes transparency, accountability, and trust between voters and representatives.

Beyond the political-science literature on manifestos, the notion of
``commitment'' in this study is also informed by speech act theory. \citet{austin1962things} distinguished performative utterances, in which
saying something also performs an action, from constative utterances, which
describe a state of affairs. He later showed that this distinction is not
always completely clear, since utterances can both describe situations and
perform actions in context.

\citet{searle1969speech, searle1976classification} further developed
this approach through the concept of illocutionary force. He classified
speech acts into assertives, directives, commissives, expressives, and
declarations. Assertives commit the speaker to the truth of a proposition,
whereas commissives commit the speaker to a future course of action, as in
promising or pledging. Declaratives are different because, under appropriate
conditions, they change an institutional or social situation when uttered. This distinction is important for our annotation scheme
(see Section~\ref{sec:data-annotation}). Although assertives also involve a type
of commitment, we use ``commitment'' in a narrower sense to refer to a
specific future action or outcome that can be evaluated as fulfilled or
unfulfilled. In our scheme, a commitment therefore identifies an actionable
and verifiable deliverable. These criteria are informed by the concept of
commissives, but specificity and verifiability are additional criteria
introduced for our annotation task.

Candidates may use commissive-like forms, such as future tense, first-person
reference, or expressions such as ``we will,'' without making a sufficiently
specific commitment. This distinction is consistent with
\citet{lamichhane2023give}'s discussion of assertive and declarative speech
acts in Nepali mayoral candidates' Facebook posts. Our error analysis
(see Section~\ref{sec:error-analysis}) also shows that both models and annotators
can be affected by the gap between surface grammatical marking and genuine
commissive content. This approach is also related to political-science research on election
pledges. The Comparative Party Pledges Project defines a pledge as ``a
statement committing a party to one specific action or outcome that can be
clearly determined to have occurred or not''
\autocite[532]{thomson2017fulfillment}. This emphasis on specific and
testable actions or outcomes is consistent with the criteria used in our
annotation scheme \autocite{naurin2019party}. Our annotation concerns whether
a statement expresses a commitment when it is made, rather than whether the
candidate later fulfills it.

Much of the literature discussed focuses on the political, sociological, and behavioral aspects of electoral commitments. But over the last decade, there also has been a growing interest in computational approaches utilizing NLP and DL to automatically analyze political discourse, manifestos, debates, and social media content. Research has explored computational analysis of political texts for tasks such as election prediction \autocite{DBLP:conf/icwsm/OConnorBRS10}, ideology classification \autocite{hirst2010partystatus}, stance detection \autocite{thomas2012votedeterminingsupportopposition}, agreement measurement \autocite{gottipati-etal-2013-learning}, and both supervised and unsupervised topic modeling, text segmentation, and text classification \autocite{article, 46934237-f0af-3c79-8cb9-4fcc1e16944e, Grimmer_2010, sevenans2014automated, menini-etal-2017-topic, glavas-etal-2016-unsupervised, Hillard15052008, Collingwood01072012, karan-etal-2016-analysis, glavas-etal-2017-cross, glavas-etal-2017-unsupervised, subramanian-etal-2018-hierarchical}. Recent studies have applied both traditional ML and advanced DL methods. \autocite{10.1007/978-3-031-74491-4_8} compares semantic classification performance between traditional ML models and DL models on Malaysian Twitter-based political texts, finding that a hybrid PSO-CNN model achieved the best results in binary classification tasks.

\citet{Fagni_2022} proposes a novel unsupervised method for inferring political leaning on social media, using deep neural network-based representation learning combined with clustering in a low-dimensional ideology space. This approach outperformed other unsupervised baselines in both fine-grained (micro $F_1$ = 0.426) and coarse-grained (micro $F_1$ = 0.772) classification tasks where the results it produced are comparable to some supervised and semi-supervised models. The study demonstrates that unsupervised representation learning with clustering offers an effective and scalable solution for political ideology inference.

Overall, these studies indicate that both supervised and unsupervised computational approaches have significant potential for advancing political text analysis, benefiting both Political Science and NLP research. In the context of Nepal,  there have been countless studies on the field of NLP and Multimodal AI such as Neural Machine Translation (NMT) \citep{chaudhary2020efforts, poudel2024bidirectional, ghimire2026neptam}, Automated Speech Recognition (ASR) \citep{ghimire2023active, ghimire2024improving, ghimire2025improving}, and Text Classification and Image/Video Captioning \citep{subedi2022cnn, subedi2023nepali, karki2026benchmarking}. However, there has been little to no prior computational research or model development on the Nepali political text mining and classification task. As a result, there is a significant research gap in the computational analysis of electoral commitments in the context of Nepal.

\section{Methods and Materials}
\label{sec:methods and materials}

\begin{figure}[h!]
	\centering
	\includegraphics[width=1.0\linewidth]{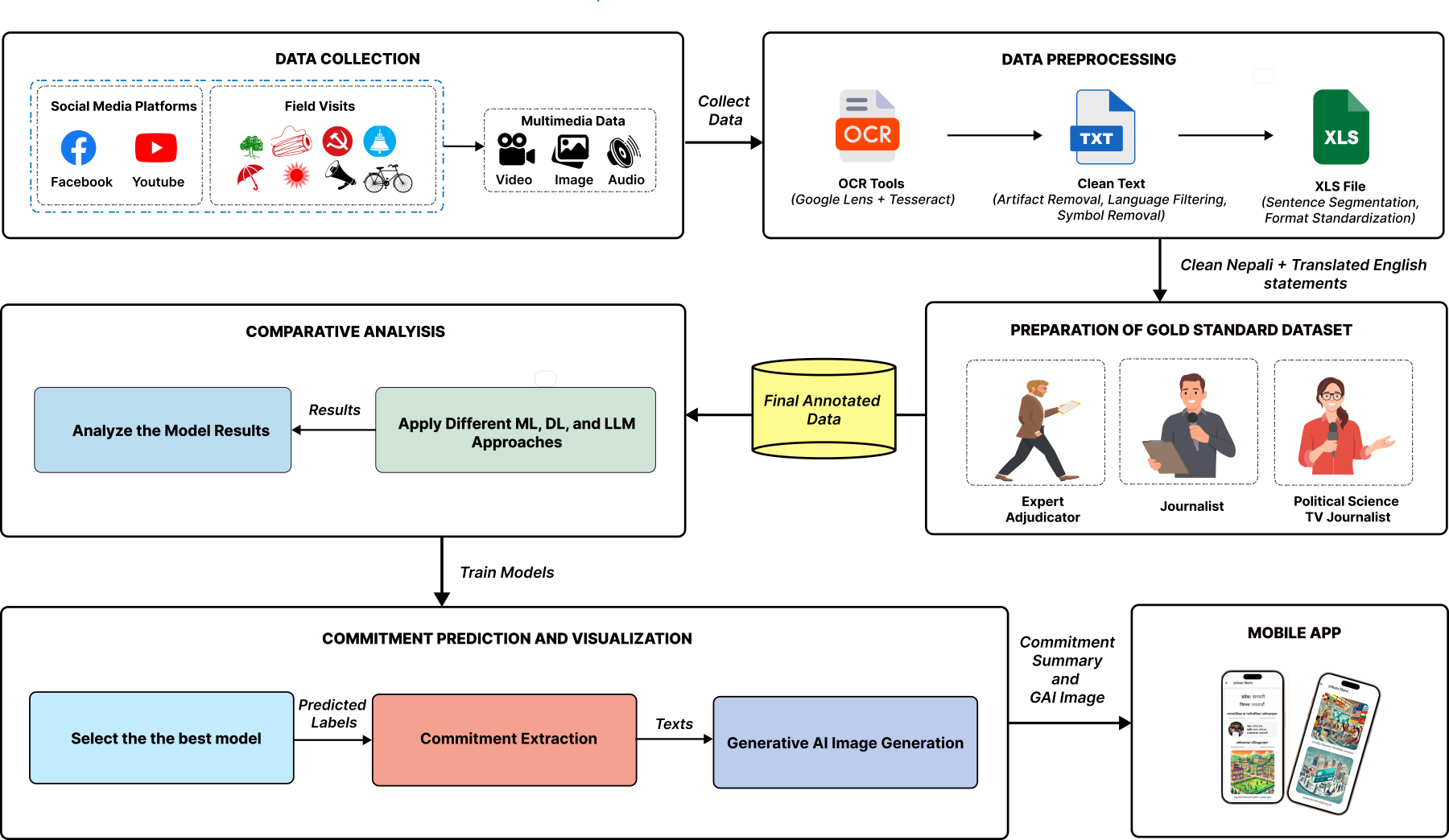}
	\caption{System architecture for mining commitments from Nepali political texts}
	\label{fig:proposed-architecture}
\end{figure}

In this section, we describe the architecture of the proposed system. The system is designed to track the steps outlined in political candidates’ pamphlets and connect these commitments to the corresponding information presented in a mobile application. The system consists of several independent modules, which are then integrated to form the complete workflow. As illustrated in Figure \ref{fig:proposed-architecture}, there are five modules, which work sequentially to fulfill the system’s requirements. The modules are as follows: data collection in Section \ref{sec:data-collection}, data preprocessing in Section \ref{sec:data-preprocessing}, and preparation of the gold standard dataset in Section \ref{sec:data-annotation}. Data distribution and analysis is explained in Section \ref{sec:data-distribution}, while the evaluation of machine learning, deep learning, large language models is presented in Section \ref{sec:ml-dl}. After extracting commitments using the best-performing model, generative AI is employed to convert the commitment texts into images, as discussed in Section \ref{sec:GAI}. Finally, these GAI-generated images, along with the commitment summaries, are integrated into a mobile application, as described in Section \ref{sec:mobile_app}.

\subsection{Data Collection}
\label{sec:data-collection}

The data is collected from five constituent bodies selected from various regions across the country. Each candidate’s election manifesto, mass meeting videos, pamphlets, and social media statements are gathered. Audio, video, and printed pamphlets are digitized and converted into text format. Data is obtained through online websites, social media platforms, and on-site field visits. Since Facebook restricts automation, we have manually collected pictures and videos of manifestos. As a result, a total of 119 images and 29 videos are collected from five provinces, as detailed in Table~\ref{tab:dataset_summary}. All data is sourced from the Facebook publicly accessible profiles of the political figures involved.
Our search has spanned 37 Nagarpalikas and has included 45 political individuals who participated in the 2022 Nepalese local-level elections. In addition, 10 publicly available manifesto documents in PDF format are collected from the official websites of various political parties.

\begin{table}[h!]
	\centering
	\caption{Dataset composition by province: candidates, images, and videos collected from Facebook profiles and party websites}
	\label{tab:dataset_summary}
	\begin{tabular}{ccccc}
		\toprule
		Provinces & Nagarpalikas & People & Images & Videos \\
		\midrule
		Koshi          & 8  & 8  & 32  & 6  \\
		\midrule
		Madhesh        & 5  & 6  & 27  & 2  \\
		\midrule
		Bagmati        & 13 & 19 & 28  & 12 \\
		\midrule
		Lumbini        & 9  & 10 & 18  & 9  \\
		\midrule
		Sudurpashchim  & 2  & 2  & 14  & 0  \\
		\midrule
		Total & 37 & 45 & 119 & 29 \\
		\bottomrule
	\end{tabular}
\end{table}

Multiple field visits are conducted to access election commitment data. Key target locations included printing presses and the offices of political parties. Among the offices of five major political parties visited, only two, Nepali Congress and the Communist Party of Nepal, provided their main party manifestos. However, they were unable to provide individual candidates’ manifestos for their respective electoral regions.

During these visits, we gain key insights indicating that such party offices do not have systems in place to archive or manage candidates’ manifestos. As a result, we additionally had to rely on other sources for data collection.

\subsection{Data Preprocessing}
\label{sec:data-preprocessing}

Before annotating the datasets for model development, the data is preprocessed which involved cleaning the data, splitting paragraphs into single sentences, and then converting them into an appropriate format for further processing. The preprocessing pipeline was implemented as a series of Python scripts utilizing the \texttt{re}\footnote{\url{https://docs.python.org/3/library/re.html}} library for regular expressions, \texttt{NumPy}\footnote{\url{https://numpy.org}} for data manipulation, \texttt{pandas}\footnote{\url{https://pandas.pydata.org}} for data handling, and \texttt{nltk}\footnote{\url{https://www.nltk.org}} for sentence segmentation adapted to Nepali text.

\begin{figure}[h!]
	\centering
	\includegraphics[width=1.0\linewidth]{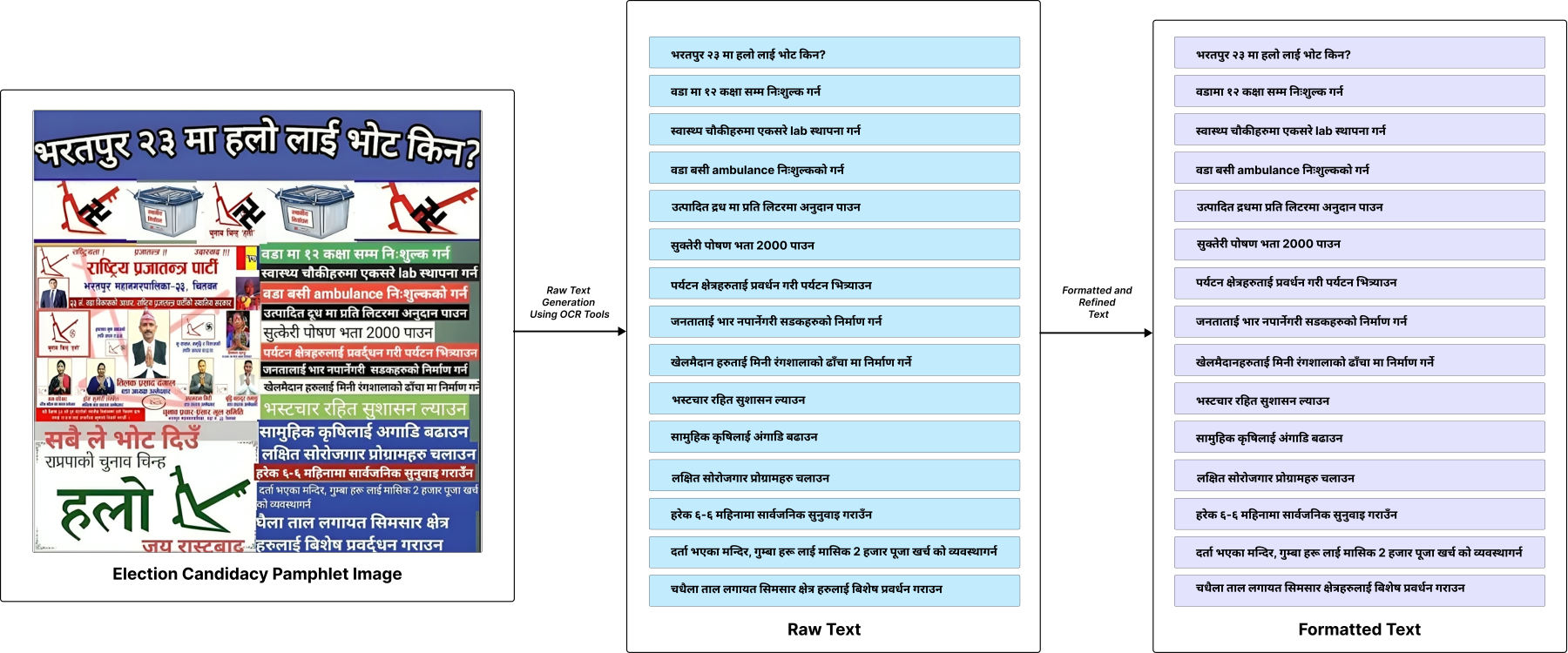}
	\caption{Preprocessing pipeline for pamphlet image-to-text conversion: from Google Lens and Tesseract OCR to refined data}
	\label{fig:data-preprocessing}
\end{figure}

As the dataset contained only Nepali text, the preprocessing pipeline, presented in Figure~\ref{fig:data-preprocessing}, involved the following steps:

\begin{itemize}[label={}]

	\item {\bfeng{Image to text conversion:}}  Transcribed the electoral commitment images, pamphlets, and posters into using Google Lens\footnote{\url{https://lens.google.com}} 
	and Tesseract OCR\footnote{\url{https://github.com/tesseract-ocr/tesseract}}.
	
	\item {\bfeng{OCR artifact removal:}} Removed duplicate whitespace, inconsistent line breaks, and OCR artifacts introduced during transcription using Python's \texttt{re} library for regular expressions.
	
	\item {\bfeng{Language filtering:}} Removed english words and non-Devanagari characters identified using Unicode range filtering by using \texttt{re} and \texttt{NumPy}, retaining only Nepali text content.

	\item {\bfeng{Symbol removal:}} Removed punctuation marks, special characters, and numeric tokens that do not contribute semantic information to commitment classification using custom regular expression patterns implemented with Python's \texttt{re} library.
	
	\item {\bfeng{Sentence segmentation:}} Split paragraphs into individual sentences using \texttt{nltk} adapted for Nepali sentence boundary markers (\nep{।} or \nep{?}) and newline characters.
	
	\item {\bfeng{Format standardization:}} Converted sentences to a standardized UTF-8 encoded format using \texttt{pandas} and exported to Excel for annotation.
\end{itemize}

Although audio and video materials were also 
collected, their processing falls outside the 
scope of the current work and is left for 
future work. So, only 28 out of 45 candidates data are processed.

\subsection{Annotation and Gold Standard Dataset Construction}
\label{sec:data-annotation}

Annotation is the most important component in the creation and construction of a gold standard dataset. In order to ensure high-quality annotations of our dataset, two bilingual annotators from diverse professional backgrounds are involved: (1) a political journalist who has expertise in legal and electoral reporting and who represents a media-oriented interpretive framework and (2) the executive director of  Nepal Television who offers insights from a policy communication and public engagement standpoint. Additionally, the final adjudication was conducted by an expert bilingual annotator-editor, a senior political journalist with over 30 years of experience covering electoral, constitutional, and political affairs in Nepal, who also has served as editor at prominent national dailies including Kantipur and The Kathmandu Post and has prior experience in reviewing both Nepali and English political documents. All annotators were selected and recruited based on their experience working with political texts and their diverse professional backgrounds and were compensated fairly based on mutual agreement between the annotators and the research team. The compensation adhered to standard remuneration practices based on the annotators' qualifications, deliverables, and the amount of time invested in the annotation process.

Initially, based on recommendations from the expert annotator, several unnecessary statements and categories, such as candidates’ names and section titles that do not fit into either the “commitment” or “non-commitment” classes, are excluded. 

\begin{table}[htbp]
	\centering
	\caption{Inter-annotator agreement confusion matrix}
	\label{tab:annotator_confusion}
	\begin{tabular}{llccc}
		\toprule
		&  & \multicolumn{3}{c}{Annotator 1 (A1)} \\[-2pt]
		\cmidrule(lr){3-5}
		& Label & C & NC & \bfeng{$N_{A1}$} \\
		\midrule
		\multirow{1}{*}{Annotator 2 (A2)} 
		& C  & 1,254 & 223  & 1,477 \\
		& NC & 44    & 909  & 953   \\
		\midrule
		& \bfeng{$N_{A2}$} & 1,298 & 1,132 & $N=2{,}430$ \\
		\bottomrule
	\end{tabular}
\end{table}

The annotation process is initiated with two domain-specific annotators, Annotator 1 and Annotator 2, both of whom are  provided with the preprocessed corpus along with comprehensive annotation guidelines. The guidelines included the following definition of an electoral commitment:

\begin{itemize}[label={}]
 \item {\bfeng{Definition:}} An electoral commitment is a statement 
made by an electoral candidate or political 
party that is specific in naming a concrete 
action or deliverable, actionable in describing 
something the candidate can realistically 
undertake if elected, and verifiable in that 
its fulfilment can be assessed after the 
election. A statement that lacks any of these 
properties, particularly specificity and 
actionability, is classified as a 
non-commitment.

\item {\bfeng{Commitment statement example:}} \nep{जलेश्वर नगरपालिका अन्तर्गत प्रत्येक वडामा नि:शुल्क} wi-fi zone \nep{व्यवस्था गर्नेछु} 
{\iteng{`I will arrange a free wi-fi zone in every ward under Jaleshwor Municipality.'}}

The above statement is classified as a commitment as it provides a concrete deliverable and scale with quantifiable noun phrase such as \nep{"नि:शुल्क"} (free) wi-fi zone and \nep{"प्रत्येक वडा"} (every ward) which describes a realistic action within an elected mandate and can be verified after the election.

\item {\bfeng{Non-commitment statement example:}} \nep{शिक्षित युवालाई आधुनिक कृषि कार्यमा लाग्न प्रेरित गरिने छ}  {\iteng{`Educated youth will be encouraged to engage in modern agricultural work.'}}

The above statement is classified as a non-commitment as it expresses a vague aspiration without any concrete or measurable outcome. Phrases such as \nep{"आधुनिक कृषि कार्य"} (modern agricultural work) and \nep{"प्रेरित गरिने"} (will be encouraged) provide a general intention rather than a specific, verifiable action, making the statement's fulfilment near impossible to assess after the election.

\end{itemize}

Based on the guidelines, each annotator independently labels the statements based on whether they reflect electoral commitments or non-commitments. Annotators apply the specificity, actionability, and verifiability criteria to distinguish vague and aspirational statements from genuine commitments. Cohen's Kappa ($\kappa$) is then calculated between these annotators to assess the annotation reliability using the values in Table \ref{tab:annotator_confusion}, shown in Equation 1.

\begin{equation}
	\kappa = \frac{P_o - P_e}{1 - P_e},
\end{equation}

where $P_o$ denotes the observed agreement between annotators and $P_e$ denotes the agreement expected by chance \autocite{cohen1960coefficient}. The value of $\kappa = 0.777$ indicates that there is a substantial level of agreement between the annotators \autocite{mchugh2012interrater}.

Later, the expert annotator conducts a final adjudication phase which involved reviewing all labeled instances independently without knowledge of which annotator assigns what label and without a prior disagreement resolution or consensus step between the two annotators. This blind adjudication enables the expert to objectively resolve inconsistencies and ambiguous cases and establish the final labels.

The Annotator reclassified 82 vague commitment statements to non-commitments that were misclassified by the annotators. For example, the statement: \nep{सबै संस्थागत विद्यालय बीच एकरूपता कायम गरिने छ र स्वस्थ प्रतिस्पर्धाको वातावरण सृजना गरिने छ} {\iteng{`Uniformity will be maintained among all institutional schools and an environment of healthy competition will be created.'}} is annotated as a commitment by both annotators. The phrase \nep{"वातावरण सृजना"} (environment will be created) suggests an outcome that lacks specific action which made it difficult to classify as a commitment.

Another such example is the statement: \nep{प्राङ्गारिक मलको प्रवर्धन र अर्गानिक खेतीको विकासलाई प्रोत्साहन गरिने छ} {\iteng{`Promotion of organic fertilizers and development of organic farming will be encouraged.'}} In this statement, the candidate uses the word \nep{"प्रोत्साहन"} (encouragement) which implied a general intention without outlining a concrete plan and also does not convey sufficient contextual clarity or commitment. Hence, this sentence is also reclassified as non-commitment.

Similarly, the commitment statements in Nepali 
were translated into English using Google 
Translate\footnote{\url{https://translate.google.com}} 
via the \texttt{GOOGLETRANSLATE} function in 
Google Sheets. The English translations were 
then reviewed by the expert annotator to verify the quality of translation. 
Both the syntactic and semantic structures of all translated statements were inspected and post-corrections were applied to 57 statements. During this process, mostly three recurring issues were identified and addressed:

 \begin{itemize}[label={}]
 	\item {\bfeng{Date and numeral/currency values:}} Currency and numeral values expressed in Western notation such as ``100,000'' and ``million'' were post-corrected to South Asian conventions such ``1 lakh'' and ``lakhs'', while date values were post-corrected to Nepali Bikram Sambat calendar conventions (``October'' corrected to ``Asoj'') to match how amounts and dates are expressed in Nepali statements.
 	
 		Nepali statement: \nep{जस अन्तरगत ५ वर्ष भित्र १ लाख नयाँ बोट रोपनि छन}
 		
 		Translated English statement: Under which 100,000 new trees will be planted within 5 years.
 		
 		Post-corrected statement: Under which 1 lakh new trees will be planted within 5 years.
 
 	\item {\bfeng{List marker error:}} Nepali statements containing numbered bullet points were mostly incorrectly translated into English where the leading numeral was interpreted as a quantity within the sentence rather than being preserved as a list marker.

 	Nepali statement: \nep{३०. पानीका स्रोत संरक्षणका लागि विशेष योजना बनाइने छ}
 	
 	Translated English statement: A special plan will be made to protect 30 water sources.
 	
 	Post-corrected statement: 30. A special plan will be made to protect water sources.

 	\item{\bfeng{Transliteration / lexical error:}}  Some translated statements contained words that were transliterated directly from Nepali instead of being translated into English. These were manually corrected so that the English sentences better reflected the original Nepali meaning and could be used reliably for image generation. In particular, where a term referred to a specific socio-historical group without a direct English equivalent (e.g., \nep{"कम्लहरी"} (Kamlahari)), the transliteration was retained, but where a component of the term had an English meaning (e.g., \nep{"मुक्त"} (freed), in \nep{"मुक्तकमैया"}), it was translated accordingly.

 	Nepali statement: \nep{मुक्तकमैया, कम्लहरी, सुकुम्बासी र अति विपन्न नागरिकहरुलाई राज्यवाट प्रदान गरिने सेवा, सुविधामा प्राथमिकता दिइनेछ}
 	
 	Translated English statement: Priority will be given to Muktakamaiyas, Kamlaharis, land squatters and extremely poor citizens in the services and facilities provided by the state.
 	
 	Post-corrected statement: Priority will be given to freed Kamaiyas, Kamlaharis, land squatters and extremely poor citizens in the services and facilities provided by the state.
 
\end{itemize}

\subsection{Data Distribution and Analysis}
\label{sec:data-distribution}

The final annotated dataset comprises a total of 2,430 instances which is categorized into two primary classes: commitment ‘C’ and non-commitment ‘NC’. Of these, 1,321 sentences are labeled as commitment and the other 1,109 are marked as non-commitment which indicates a slightly skewed distribution. Additional analysis is conducted on the dataset to gain further insights such as the distribution of data across each province which is presented in Table \ref{tab:data_distribution}. The table shows that no data was collected from Province 4 (Gandaki) and Province 6 (Karnali). The highest number of instances is from Province 3 (Bagmati).

\begin{table}[t!]
	\centering
	\caption{Distribution of commitment and non-commitment statements by province}
	\label{tab:data_distribution}
	\begin{tabular}{cccc}
		\toprule
		Provinces & No. of Candidates/Parties & No. of Commitments (C) & No. of Non-Commitments (NC) \\
		\midrule
		Koshi            & 6  & 228 & 99 \\
		\midrule
		Madhesh          & 6  & 196 & 64  \\
		\midrule
		Bagmati          & 11 & 688 & 899 \\
		\midrule
		Lumbini          & 3  & 108 & 30  \\
		\midrule
		Sudurpashchim    & 2  & 101 & 17  \\
		\midrule
		Total   & 28 & 1,321 & 1,109 \\
		\bottomrule
	\end{tabular}
\end{table}

\subsection{Evaluated Models}
\label{sec:ml-dl}
\subsubsection{Machine Learning Models}

For binary classification of Nepali political 
commitment texts, we benchmarked six classical 
machine learning algorithms. \texttt{Logistic Regression} (LR) 
estimates the probability of class membership 
by fitting data to a logistic curve 
\autocite{park2013introduction}.  \texttt{Decision tree} (DT) is a flow-chart-like tree structure where each internal node is denoted by rectangles and the leaf nodes are denoted by ovals \autocite{priyam2013comparative} and is widely used because of its ease of implementation and interpretability compared to other algorithms \autocite{yadav2012data}. \texttt{Random Forest} (RF) aggregates multiple tree classifiers to improve prediction accuracy and reduce 
overfitting \autocite{liu2012new}. \texttt{Support Vector 
Machine} (SVM) builds optimal decision boundaries 
from limited samples \autocite{Vapnik_1995}. 
\texttt{Naïve Bayes} (NB) computes the probability of belonging to a class assuming conditional independence between attributes \autocite{langley1994induction}. Finally, the \texttt{K-Nearest Neighbor} (KNN) algorithm assigns an unclassified sample to the class of its nearest labeled neighbor \autocite{cover1967nearest}. 

We perform a comparative analysis for each model under two text 
representations, Term Frequency–Inverse Document Frequency (TF-IDF) \autocite{salton1988term} and pretrained Language-agnostic BERT Sentence Embedding (LaBSE) \autocite{feng2022language} embeddings to assess 
the impact of embedding quality on classification 
performance. 

\subsubsection{Deep Learning Models}

In order to capture sequential and semantic patterns beyond that of classical models, we also employed a range of deep learning architectures.  \texttt{Simple Recurrent Neural Networks} (SRNs) \autocite{elman1990finding} capture temporal dependencies by transferring hidden layer activations to a context layer at each time step which serves as a model for sequence modeling in Nepali text. \texttt{Long Short-Term Memory} (LSTM) networks \citep{hochreiter1997long} address the vanishing gradient problem of SRNs \autocite{hochreiter1991untersuchungen, hochreiter2001gradient} through specialized gating mechanisms that enable the modeling of long-term dependencies. Similarly, \texttt{Gated Recurrent Units} (GRUs) \autocite{cho2014properties} which simplify LSTM design by omitting explicit memory cells while retaining gating mechanisms \autocite{chung2014empirical} are applied for their efficiency in learning linguistic dependencies with fewer parameters. One-Dimensional \texttt{Convolutional Neural Networks} (CNNs) \citep{kim2014convolutional} have shown strong performance on limited labeled datasets by detecting position-invariant local text patterns \autocite{kiranyaz20211d}, and in our work are used to extract n-gram features from Nepali text.
All Recurrent Neural Network (RNN) and CNN models were evaluated with 
both fastText \autocite{bojanowski2017enriching} 
and GloVe \citep{pennington2014glove} embeddings trained on the political 
commitment corpus. \texttt{Multilingual BERT} (mBERT) 
\autocite{devlin2019bert} and \texttt{XLM-RoBERTa} 
(XLM-R) \autocite{conneau2019unsupervised} 
were fine-tuned on our dataset as transformer 
models. 

\subsubsection{Large Language Models}

In order to contextualize fine-tuned model performance 
against modern general-purpose causal language models, 
we evaluated three LLMs in zero-shot and 
few-shot settings without any task-specific 
fine-tuning: \texttt{Qwen3-8B} \autocite{qwen3technicalreport}, 
\texttt{Llama-3.1-8B-Instruct} \autocite{grattafiori2024llama}, 
and \texttt{Gemma-4-E4B-it} \autocite{gemmateam2026gemma4}. 
In the zero-shot setting, models were prompted 
with the commitment definition and asked to 
classify each statement without any examples. 
In the few-shot setting, five representative 
examples, five commitments and five 
non-commitments drawn from the training set, were provided as in-context demonstrations 
\autocite{brown2020language}. This evaluation 
allows us to assess whether domain-specific 
fine-tuning on our gold-standard dataset 
provides meaningful advantages over 
general-purpose LLMs without task-specific 
training and also allows us to access its performance against fine-tuned DL models.

\subsection{Generative AI for Visual Representation of Classified Commitments}
\label{sec:GAI}
In this section, we discuss the application of GAI for text to image conversion based on the commitments extracted by our best performing model identified in the experiments and results section. Rather than using web-sourced images, we use GAI-generated images, as GAI has become the disruptor in the digital environment \autocite{dwivedi2023opinion, wessel2025generative} that can produce high-quality, contextually relevant content that is difficult to distinguish from human work. Specifically, we use Stable Diffusion \autocite{rombach2022high} to generate images from text labelled as `Commitment'.

From the dataset, a subset of commitments labelled `C' was extracted. These commitments were originally in Nepali and were translated into English using the Google Translate API. The translated text was then used to build image prompts, with the prompt design focused on contextual relevance and visual interpretability so that each prompt captured the underlying commitment.

Two methods were used to generate images with Stable Diffusion:

\begin{itemize}
	\item {\bfeng{Local inference}} using the Hugging Face \texttt{diffusers} library, running the CompVis/stable-diffusion-v1-4 model\footnote{\url{https://huggingface.co/CompVis/stable-diffusion-v1-4}} via a Python script with GPU acceleration where available. Parameters such as guidance scale, inference steps, and prompt formulation were tuned to control image quality and alignment.
		\item {\bfeng{Web interface automation}} using Selenium, where a headless browser loads an online implementation of Stable Diffusion, inserts the formatted prompt, triggers generation, and retrieves the resulting image. This avoids running the model locally, and reduces the computational cost, but comes with two limitation: (1) limited access to main parameters (inference steps, resolution, seed values) and (2) dependency on the stability and availability of the third-party website.

	Despite this, the method worked well for rapid prototyping and in cases where local compute was limited.
\end{itemize}

\begin{figure}[htp]
	\centering
	\begin{subfigure}[b]{0.49\linewidth}
		\centering
		\includegraphics[height=5cm, keepaspectratio]{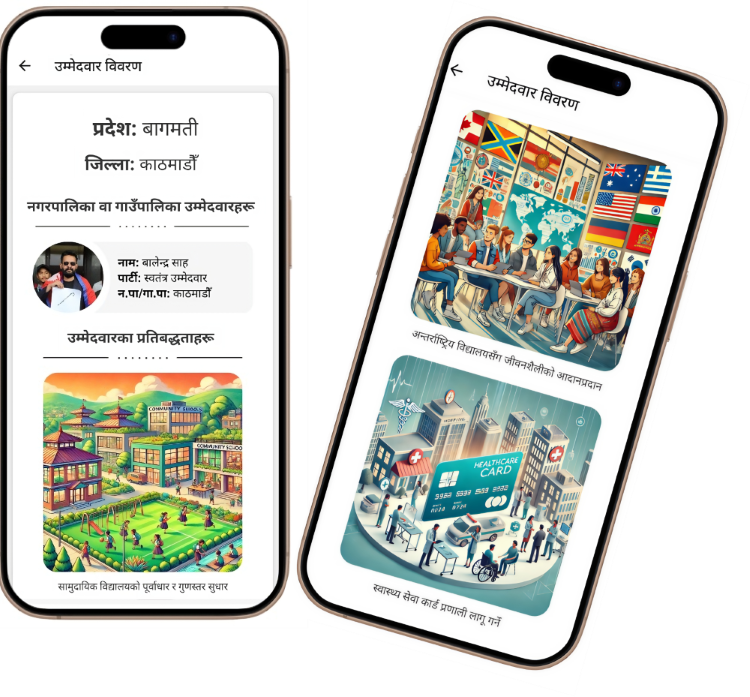}
		\caption{Balendra Shah}
		\label{fig:balendra-shah}
	\end{subfigure}
	\hfill
	\begin{subfigure}[b]{0.49\linewidth}
		\centering
		\includegraphics[height=5cm, keepaspectratio]{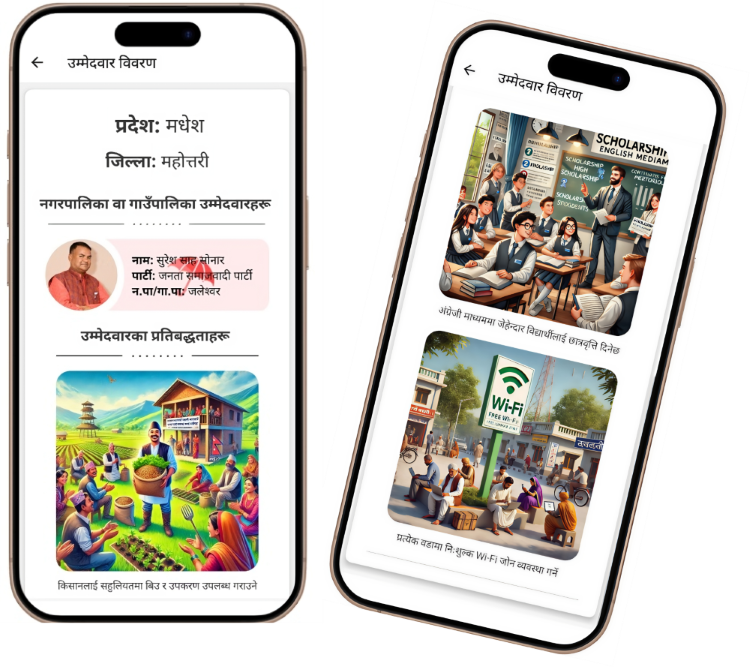}
		\caption{Suresh Sah Sonar}
		\label{fig:suresh-sonar}
	\end{subfigure}
	\caption{Profile images of the candidates as displayed in the mobile application}
	\label{fig:candidates}
\end{figure}

For both methods, generated images were checked for how well they reflected the translated commitments, and prompts were revised iteratively where needed to improve semantic and visual alignment. We used the second approach for this study, as it produced more detailed and representative images in less time. A few examples of images generated for electoral commitments are shown in Figure \ref{fig:candidates}. 

The generated images were reviewed internally 
by the research team and workshop evaluation 
committee to assess their visual and semantic 
alignment with the source commitment statements. 
For the project, 39 images were generated 
for 11 selected candidates, representing all 
five provinces. Reviewers evaluated each image 
to determine whether it reflected the core 
theme of the corresponding commitment without 
introducing unrelated content.

Overall, the generated images aligned accurately
with the core aspect of their corresponding 
commitments. However, one recurring limitation 
was observed for commitments with limited 
contextual (geographic and demographics) information. For example, a commitment such as  {\iteng{`New roads will be 
constructed and old roads repaired and 
upgraded as needed'.}} We can see in this commitment, no geographic context such as the municipality, ward, or geographic location is mentioned. So, the model tends to generate images showcasing Western 
road systems, which is misleading for 
Nepali users. To address this, additional 
contextual information about the location 
of the commitment was incorporated into 
the image generation prompt through the candidates’ metadata like electoral constituency, which improved the visual relevance of the generated images.

We acknowledge that this internal review does not meet the standards of a formal user study or expert evaluation. The purpose of this review is to confirm that the generated images function as illustrative visual representations meant to enhance public engagement with candidate commitments. They should not be interpreted as factual depictions of actual events or outcomes.

\subsection{Mobile Application Development}
\label{sec:mobile_app}

In this section, we discuss on how we integrate a mobile application (functional prototype) to display AI-generated images with summarized commitment texts for subset of candidates, representing all five provinces. The app consists of two components: one for rendering the map of Nepal with specific demarcations (i.e., Province and District), and the other for displaying details of candidates (i.e., Candidates’ name, affiliated political parties, electoral constituency, and manifestos/commitments) from specific districts based on commands. The actual commitments from candidates manifestos or pamphlets are summarized to offer users a concise overview of each candidate’s promises.

The mobile application plays a crucial role in bridging the gap between electoral commitments and public awareness. It offers an accessible digital platform through which users can explore and evaluate candidates' commitments and performance in a visual and interactive manner. Given that, according to the National Population and Housing Census 2021 National Report\footnote{\url{https://censusresults.nsonepal.gov.np/household}}, 73\% of households in Nepal own smart mobile phones and 37.78\% have access to the internet. As a result, we are trying to leverage widespread mobile access to increase civic engagement. By delivering candidate information directly to the public, the app helps constituents hold elected officials accountable and make informed decisions in future elections (i.e., either to re-elect the same candidate in the next election or not).

\begin{figure}[htp]
	\centering
	\includegraphics[width=0.7\linewidth]{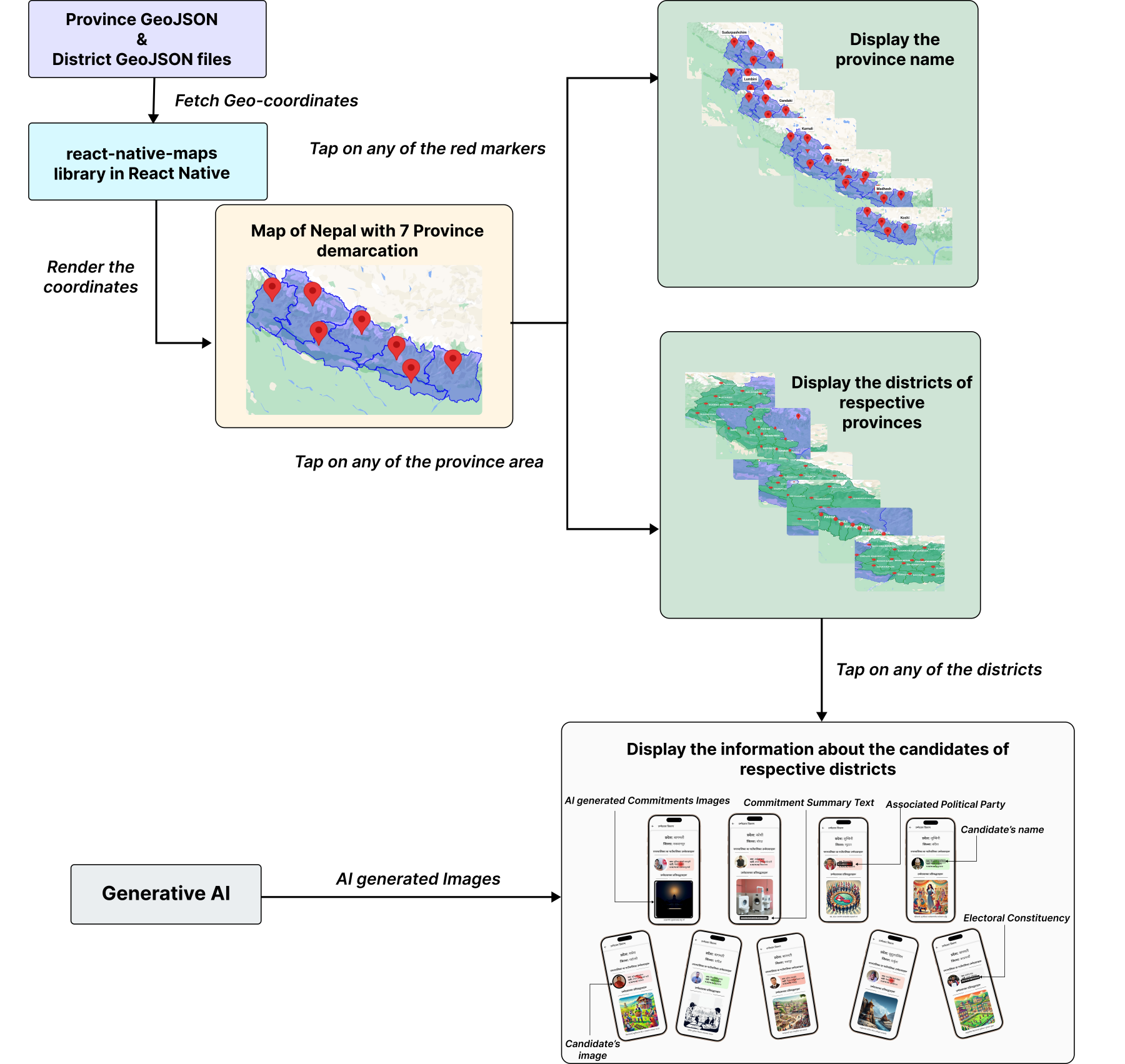}
	\caption{Workflow diagram of the mobile application}
	\label{fig:mobile-app-architecture}
\end{figure}

The workflow diagram of the mobile app is shown in Figure \ref{fig:mobile-app-architecture} which demonstrates the process, starting with the rendered map of Nepal, showing how users can interact with the app, from exploring provinces and districts on the map to viewing candidates’ manifesto images from Generative AI along with the summary commitments texts for different electoral constituencies. Some of the examples are represented in Table \ref{tab:candidate_commitments}.

\begin{table}[htp]
	\caption{A few example candidate commitments shown in original Nepali, summarized Nepali, and English translation}
	\label{tab:candidate_commitments}
	\centering
	\begin{tabular}{l p{3.5cm} p{3cm} p{3.5cm}}
		\toprule
		Name of Candidate & Commitments (in Nepali) & Summarized Commitments (in Nepali) & Commitments (in English) \\
		\midrule[0.5pt]
		\nep{बालेन्द्र शाह} [Figure \ref{fig:balendra-shah}] & 
		\nep{काठमाडौं महानगरपालिकाका सामुदायिक विद्यालयहरूको पूर्वाधार, गुणस्तर र वातावरणमा सुधार गर्नु} & 
		\nep{सामुदायिक विद्यालयको गुणस्तर सुधार} & 
		To improve the infrastructure, quality, and environment of community schools in Kathmandu Metropolitan City and to emphasize technical education. \\
		
		& \nep{अन्तर्राष्ट्रिय विद्यालयहरूसँगको सहकार्य गरी जीवनशैली, संस्कृति, शिक्षा प्रणाली आदिको आदानप्रदान गर्नु} & 
		\nep{अन्तर्राष्ट्रिय विद्यालयसँग जीवनशैलीको आदानप्रदान} & 
		International schools will be partnered with national schools and sisterhood relationships will be initiated to exchange culture, lifestyle, education systems, etc. \\
		
		\midrule
		
		\nep{सुरेश साह सोनार} [Figure \ref{fig:suresh-sonar}] &
		\nep{कृषि क्षेत्रको प्रवर्द्धन एवं विकासको लागि किसानलाई सहज तथा समयमा सहुलियत रकममा बिउ, बिजन र कृषि उपकरण उपलब्ध गराउने छु} &
		\nep{किसानलाई सहुलियतमा बिउ र उपकरण उपलब्ध गराउने} &
		For the promotion and development of the agricultural sector, I will provide seeds, seedlings and agricultural equipment to farmers easily and on time at subsidized rates. \\
		
		& \nep{प्रत्येक सामुदायिक विद्यालयमा अंग्रेजी माध्यमबाट पठन पाठन संचालन गर्नको साथै उच्च-शिक्षा हासिल गर्न जेहेन्दार छात्र छात्राहरुलाई छात्रवृत्ति प्रदान गर्ने छु} &
		\nep{अंग्रेजी माध्यममा जेहेन्दार विद्यार्थीलाई छात्रवृत्ति दिनेछ} &
		I will provide scholarships to meritorious students to pursue higher education by conducting English medium teaching in every community school. \\		
	
		\bottomrule
	\end{tabular}
\end{table}

\section{Experiments and Results}
\label{sec:experiments_results}

The experimental section presents the comparative evaluation of the proposed models using the annotated dataset. This section outlines six subsections: Section~\ref{sec:experiment-setup} describes the experimental setup, Section~\ref{sec:evaluation-metrics} presents the evaluation metrics,
Section\ref{sec:statistical-testing} outlines the statistical significance test used, Section~\ref{sec:results} reports the results and findings of the experiments, Section~\ref{sec:ablation-study} presents two ablation studies, and Section~\ref{sec:error-analysis} presents an error analysis of the best-performing models.

\subsection{Experimental Setup}
\label{sec:experiment-setup}
In this section, we describe the experimental setup for our experiments. We first utilize the gold standard dataset which consists of 2,430 instances labeled as either commitment (C) or non-commitment (NC). Among these, 1,321 are C and the remaining 1,109 are NC. We stratify the dataset into an 80-20 train-test split. The stratification is done based on four attributes: province, district/gaupalika, sentence length, and final label. The 20\% (486 instances) test data is held out for final evaluation purposes, and the remaining 80\% is used for training. It consists of 1,050 C sentences and 894 NC sentences. Due to class imbalance in the NC class, 155 new instances are augmented using three techniques (i.e., random word swap, random deletion, and random insertion) in the training data, resulting in 1,049 NC instances. All instances are then reviewed by reviewers to ensure that there is no data leakage in the test set. For model training, we apply 5-fold cross-validation on the 80\% split for all ML and DL models. The choice of 5-fold cross-validation follows the recommendation in \autocite{raschka2018model}, which states that k-fold cross-validation is the best available non-parametric method for assessing model generalization, and the advice in \autocite{arlot2016choice} to use k=5, given our computational constraints.

We perform a comparative analysis between traditional ML models, RNN and CNN architectures, two fine-tuned BERT-base models (i.e., mBERT and XLM-R), and general-purpose Multilingual LLMs to determine which model performs best on the political text classification task. In the context of embedding of DL models, we domain-adapted fastText embeddings by initializing them from pretrained Nepali vectors and further training them on the electoral commitment training dataset using the skip-gram model with embedding dimension of 300, 20 epochs, learning rate (lr) of 0.05, bigram features (wordNgrams=2), minimum word count of 3, and context window size of 5. GloVe embeddings were similarly trained in the same training corpus with embedding dimension of 300, lr of 0.05, 100 epochs, minimum word count of 1, and context window size of 5. For model configuration, the hyperparameters for ML models are selected through grid search based on the average fold macro $F_1$-score, whereas for DL models, the hyperparameters are selected through random search with 20 trials, following \autocite{bergstra2012random}, who show that random search finds models comparable to grid search within a small fraction of the computation time, making it a more practical choice for training higher-parameter models under our computational constraints. The final model is trained on the 80\% train data using the optimized hyperparameter set. For DL models, the same 80\% data is further split into 90-10 train-validation set. The validation set is used for early stopping with patience=3. The best model configuration for fastText-LSTM is lr of \begin{math} 3e-4\end{math}, 30 epochs, 1 LSTM layer (rnn\_units=32), a dropout rate of 0.3, a weight decay of 0.0005, and a batch size of 16. Similarly, the complete hyperparameter configurations for all evaluated models are provided in Section~\ref{sec:appendix_hyperparams}.

We conduct the experiments on a workstation with an Intel(R) Xeon(R) Platinum 8255C CPU @ 2.50GHz and a GeForce RTX 4060 Ti 16GB, utilizing scikit-learn v1.9.0 for ML models, PyTorch v2.12.1 for DL models, and huggingface hub v1.24.0, transformers v5.14 for LLMs, with Python 3.13.11 as the programming language. We use the Mersenne Twister pseudorandom number generator \autocite{matsumoto1998mersenne} for NumPy and the Philox counter-based random number generator \autocite{salmon2011parallel} for PyTorch. All experiments are executed with a fixed random seed of 42 to ensure reproducibility.

\subsection{Evaluation Metrics}
\label{sec:evaluation-metrics}
Several performance measures are used to give a complete evaluation of the models which cover different aspects of how accurately they make predictions. Using multiple measures allows for a thorough assessment and a clearer understanding of each model’s strengths and weaknesses. The measures used include accuracy, macro recall (ma-Recall), macro precision (ma-Precision), macro $F_1$-score (ma-$F_1$-score), and the area under the receiver operating characteristic curve (AUROC).

\subsection{Statistical Significance Testing}
\label{sec:statistical-testing}

In this section, we discuss the statistical significance test employed in our experiments to verify whether the observed performance differences between the models genuinely reflect differences in their performance or are simply due to sampling variation. There are several suitable parametric and non-parametric statistical tests used in NLP research, including the paired Student's $t$-test \autocite{fisher1937design}, Sign test \autocite{gibbons2010nonparametric}, McNemar's test \autocite{mcnemar1947note}, Cochran's $Q$ test \autocite{cochran1950comparison}, Wilcoxon signed-rank test \autocite{wilcoxon1945individual}, Pitman's permutation test \autocite{noreen1989computer}, and Paired bootstrap test \autocite{efron1994introduction}. We select McNemar's test because it best suits our experimental setting, as it is designed for paired nominal observations (binary labels), is appropriate for binary classification tasks, and has been widely used in NLP research \autocite{dror2018hitchhiker, blitzer2006domain, ziser2017neural}. This test is applied to a $2 \times 2$ contingency table that summarizes the paired predictions of two models on the same set of test instances (486 instances in our case). There are four versions of McNemar's test: the classical test, continuity-corrected test, exact test, and mid-$P$ test. We opt for the McNemar mid-$P$ test because it is useful in our empirical experiments where we are working with small and moderate numbers of discordant pairs, and also it strikes a balance between the overly conservative exact test and asymptotic methods that may frequently deviate from the nominal significance level ($\alpha = 0.05$ in our case). 

We test the following hypotheses regarding the marginal performance of Model~A and Model~B:

\begin{align}
	H_0&: \quad \text{Model A and Model B have equal performance} \quad (\pi = 0.5) \\
	H_1&: \quad \parbox[t]{10cm}{Model A and Model B do \emph{not} have equal
	performance, i.e., one model performs better than the other} \quad (\pi \neq 0.5)
\end{align}

where $\pi = P(\text{a discordant pair favors Model A})$. This is a \emph{two-tailed} test: $H_1$ does not specify \emph{which} model is better; it only specifies that the two models' error rates differ.

Let $b$ and $c$ denote the numbers of discordant pairs in a $2 \times 2$ contingency table, where $b$ is the count of cases where Model~A is correct and Model~B is wrong, and $c$ is the count where Model~A is wrong and Model~B is correct. Define $n = b + c$. Under $H_0$, the number of discordant pairs favoring Model~A follows a binomial distribution, $b \sim \text{Binomial}(n, 0.5)$. Without loss of generality, assume $b \le c$ (otherwise swap $b$ and $c$), so that $b$ is the smaller discordant count.

The two-sided exact McNemar $p$-value is then

\begin{equation}
	p_{\text{exact}} 
	= 2 \sum_{i=0}^{b} \binom{n}{i} 0.5^{n}.
\end{equation}

The mid-$p$ value is obtained by subtracting the probability mass at the observed configuration:

\begin{equation}
	p_{\text{mid-p}} 
	= p_{\text{exact}} - \binom{n}{b} 0.5^{n}
	= 2 \sum_{i=0}^{b} \binom{n}{i} 0.5^{n} \;-\; \binom{n}{b} 0.5^{n}.
\end{equation}

We fix the significance level at $\alpha = 0.05$ and reject $H_0$ in favor of $H_1$ if

\begin{equation}
	p_{\text{mid-p}} < \alpha = 0.05.
\end{equation}

If $p_{\text{mid-p}} \geq 0.05$, we fail to reject $H_0$, and conclude there is insufficient evidence that either model outperforms the other.

\subsection{Results}
\label{sec:results}

We report results for classical machine learning models, deep learning architectures, and large language models evaluated in zero-shot and few-shot settings as listed in Table~\ref{tab:ml_model_performance}, Table~\ref{tab:dl_model_performance}, and Table~\ref{tab:mllm_model_performance} respectively. For ML models, LaBSE embeddings showed a slight performance advantage over classical TF-IDF across all model families. LaBSE-LR achieved the best performance with a macro $F_1$-score of 0.896, followed closely by LaBSE-SVM at macro $F_1$-score of 0.892. For deep learning models, fastText embeddings showed higher performance than GloVe across all architectures. Among RNN-based architectures, all models had comparable results with fastText-LSTM edged over fastText-GRU (0.904 macro $F_1$-score) and fastText-SRN (0.898 macro $F_1$-score) with a macro $F_1$-score of 0.913 and AUROC of 0.961. fastText-LSTM also achieved higher performance than all ML, CNN, and transformer-based models evaluated. For transformer architectures, under identical fine-tuning configurations, XLM-R and mBERT achieved comparable performance (marginal difference) with macro $F_1$-scores of 0.882 and 0.875 respectively. For large language models, few-shot prompting showed higher performance than zero-shot across all evaluated models. Qwen3-8B achieved the best performance among LLMs with a macro $F_1$-score of 0.848, followed by Llama-3.1-8B-Instruct at 0.841 macro $F_1$-score. However, compared to a fine-tuned model such as fastText-LSTM with far less parameters, both fell short with fastText-LSTM model achieving a +6.50\% higher macro $F_1$-score than the best performing LLM model, Qwen3.

\begin{table}[H]
	\centering
	\caption{Performance of ML models under TF-IDF and LaBSE text representations. The best result is \bfeng{bolded}, and the second best is \underline{underlined}.}
	\label{tab:ml_model_performance}
	\begin{tabular}{l  c c c c c}
		\toprule
		Model Name & Accuracy & 
		ma-Precision  & 
		ma-Recall & 
		ma-$F_1$-score & 
		AUROC \\
		\midrule
		TF-IDF-LR & 0.870 & 0.869 & 0.869 & 0.869 & 0.943 \\
		LaBSE-LR  & \bfeng{0.897} & \bfeng{0.896} & \bfeng{0.896} & \bfeng{0.896} & \underline{0.947} \\
		\midrule
		TF-IDF-DT & 0.735 & 0.736 & 0.740 & 0.735 & 0.809 \\
		LaBSE-DT  & 0.747 & 0.744 & 0.741 & 0.742 & 0.813 \\
		\midrule
		TF-IDF-RF & 0.860 & 0.858 & 0.861 & 0.859 & 0.931 \\
		LaBSE-RF  & 0.870 & 0.871 & 0.866 & 0.868 & 0.939 \\
		\midrule
		TF-IDF-SVM & 0.870 & 0.869 & 0.869 & 0.869 & 0.945 \\
		LaBSE-SVM  & \underline{0.893} & \underline{0.891} & \underline{0.893} & \underline{0.892} & \bfeng{0.952} \\
		\midrule
		TF-IDF-NB & 0.865 & 0.870 & 0.868 & 0.862 & 0.925 \\
		LaBSE-NB  & 0.875 & 0.875 & 0.871 & 0.872 & 0.927 \\
		\midrule
		TF-IDF-KNN & 0.856 & 0.862 & 0.848 & 0.852 & 0.927 \\
		LaBSE-KNN  & 0.862 & 0.864 & 0.856 & 0.859 & 0.925 \\
		\bottomrule
	\end{tabular}
\end{table}

\begin{table}[H]
	\centering
	\caption{Performance of DL models under fastText and GloVe embeddings, and transformer-based models. The best result is \bfeng{bolded}, and the second best is \underline{underlined}.}
	\label{tab:dl_model_performance}
	\begin{tabular}{l c c c c c}
		\toprule
		Model Name & Accuracy & 
		ma-Precision  & 
		ma-Recall & 
		ma-$F_1$-score & 
		AUROC \\
		\midrule
		
		fastText-SRN & 0.899 & 0.898 & 0.899 & 0.898 & 0.951 \\
		GloVe-SRN   & 0.852 & 0.850 & 0.854 & 0.851 & 0.924 \\
		\midrule
		fastText-LSTM  & \bfeng{0.914} & \bfeng{0.912} & \bfeng{0.915} & \bfeng{0.913} & \bfeng{0.961} \\
		GloVe-LSTM   & 0.840 & 0.841 & 0.845 & 0.839 & 0.915 \\
		\midrule
		fastText-GRU   & \underline{0.905} & \underline{0.904} & \underline{0.904}  & \underline{0.904} & \underline{0.959} \\
		GloVe-GRU   & 0.850 & 0.849 & 0.853 & 0.849 & 0.920 \\
		\midrule
		fastText-CNN     & 0.889 & 0.887 & 0.890 & 0.888 & 0.959 \\
		GloVe-CNN   & 0.858 & 0.856 & 0.860 & 0.857 & 0.933 \\
		\midrule
		XLM-R       & 0.883 & 0.881 & 0.886 & 0.882 & 0.940 \\
		mBERT & 0.877 & 0.876 & 0.873 & 0.875 & 0.931 \\
		\bottomrule
	\end{tabular}
\end{table}

\begin{table}[H]
	\centering
	\caption{Performance of LLMs under zero-shot and few-shot prompting. The best result is \bfeng{bolded}, and the second best is \underline{underlined}.}
	\label{tab:mllm_model_performance}
	\begin{tabular}{l c c c c c}
		\toprule
		Model Name & Accuracy & 
		ma-Precision  & 
		ma-Recall & 
		ma-$F_1$-score & 
		AUROC \\
		\midrule
		Gemma-4-E4B-it-\iteng{zero-shot} & 0.815 & 0.819 & 0.822 & 0.815 & 0.822 \\
		Gemma-4-E4B-it-\iteng{few-shot} & 0.833 & 0.838 & 0.841 & 0.833 & 0.841 \\
		
		\midrule
		Llama-3.1-8B-Instruct-\iteng{zero-shot} & 0.609 & 0.687 & 0.639 & 0.593 & 0.639 \\
		Llama-3.1-8B-Instruct-\iteng{few-shot} & \underline{0.842} & \underline{0.844} & \underline{0.848} & \underline{0.841} & \underline{0.848} \\
		\midrule
		Qwen3-8B-\iteng{zero-shot} & 0.817 & 0.831 & 0.829 & 0.817 & 0.829 \\
		Qwen3-8B-\iteng{few-shot} & \bfeng{0.848} & \bfeng{0.855} & \bfeng{0.857} & \bfeng{0.848} & \bfeng{0.857} \\
		
		\bottomrule
	\end{tabular}
\end{table}

To validate these findings, we assessed their statistical significance using McNemar's mid-$p$ test as discussed in Section~\ref{sec:statistical-testing}. Table~\ref{tab:mcnemar_results} reports pairwise test results across the model comparisons. In the case where the result is statistically significant, the direction of superiority is determined by $c > b$ which indicates that the Model B performs better than Model A (the higher-performing model is placed in the Model B column of Table~\ref{tab:mcnemar_results}).

Among RNN architectures, fastText-LSTM showed no statistically significant difference against fastText-SRN and fastText-GRU ($p$ = 0.248 and $p$ = 0.572 respectively). This suggests that the three RNN variants perform equivalently for our task. However, fastText-LSTM showed a statistically significant advantage over fastText-CNN ($p$ = 0.0227). Similarly, fastText-LSTM demonstrated a statistically significant edge over XLM-R ($p$ = 0.0167), TF-IDF-LR, and TF-IDF-SVM ($p$ = 0.00145), while no significant difference was found against LaBSE-LR ($p$ = 0.200) and LaBSE-SVM ($p$ = 0.121), indicating that the best DL and ML models achieve comparable performance on this task. fastText embeddings also showed a statistically significant improvement over GloVe across both LSTM ($p$ = 0.000002) and CNN ($p$ = 0.0402) architectures. No significant difference was found between mBERT and XLM-R ($p$ = 0.659). For large language models, few-shot prompting showed a statistically significant improvement over zero-shot for Qwen3-8B ($p$ = 0.0113), while Llama-3.1-8B-Instruct and Qwen3-8B performed comparably in the few-shot setting ($p$ = 0.683). fastText-LSTM demonstrated a statistically significant advantage over the best zero-shot LLM ($p$ = 0.000042). Finally, while LaBSE edged over TF-IDF numerically, this difference did not reach statistical significance for best performing ML model, (TF-IDF-LR vs LaBSE-LR, $p$ = 0.0704), and also no significant difference was found between the two best performing ML models, LaBSE-LR and LaBSE-SVM ($p$ = 0.508).

\begin{table}[H]
	\centering
	\caption{Pairwise McNemar mid-$p$ test results. Model~B denotes the
		higher-performing model (higher $F_1$-score) in each comparison. $b$ = pairs where Model~A is
		correct and Model~B is wrong; $c$ = pairs where Model~A is wrong and
		Model~B is correct; $n = b + c$. The statistically significant pairs are \bfeng{bolded}.}
	\label{tab:mcnemar_results}
	\begin{tabular}{l l c c c c }
		\toprule
		Model A & Model B & $b$ & $c$ & $n$ & $p_{\text{mid-}p}$ \\
		\midrule
		fastText-SRN & fastText-LSTM & 10 & 16 & 26 & 0.248 \\
		\midrule
		fastText-GRU & fastText-LSTM & 12 & 15 & 27 & 0.572 \\
		\midrule
		\bfeng{fastText-CNN} & \bfeng{fastText-LSTM} & \bfeng{6} & \bfeng{17} & \bfeng{23} & \bfeng{0.0227} \\
		\midrule
		\bfeng{XLM-R}  & \bfeng{fastText-LSTM} & \bfeng{10} & \bfeng{24} & \bfeng{34} & \bfeng{0.0167} \\
		\midrule
		\bfeng{TF-IDF-LR}    & \bfeng{fastText-LSTM} & \bfeng{10} & \bfeng{30} & \bfeng{40} & \bfeng{0.00145} \\
		\midrule
		\bfeng{TF-IDF-SVM}    & \bfeng{fastText-LSTM} & \bfeng{10} & \bfeng{30} & \bfeng{40} & \bfeng{0.00145} \\
		\midrule
		\bfeng{Qwen3-8B (zero-shot)} & \bfeng{fastText-LSTM} & \bfeng{14} & \bfeng{45} & \bfeng{59} & \bfeng{0.000042} \\
		\midrule
		LaBSE-LR     & fastText-LSTM & 11 & 18 & 29 & 0.200 \\
		\midrule
		LaBSE-SVM    & fastText-LSTM & 12 & 21 & 33 & 0.121 \\
		\midrule
		mBERT        & XLM-R   & 21 & 24 & 45 & 0.659 \\
		\midrule
		\bfeng{GloVe-LSTM} & 	\bfeng{fastText-LSTM} & 	\bfeng{11} & 	\bfeng{46} & 	\bfeng{57} & 	\bfeng{0.000002} \\
		\midrule
		\bfeng{Glove-CNN} & \bfeng{fastText-CNN} & \bfeng{19} & \bfeng{34} & \bfeng{53} & \bfeng{0.0402} \\
		\midrule
		\bfeng{Qwen3-8B (zero-shot)} & \bfeng{Qwen3-8B (few-shot)} & \bfeng{10} & \bfeng{25} & \bfeng{35} & \bfeng{0.0113} \\
		\midrule
		Llama-3.1-8B-Instruct (few-shot) & Qwen3-8B (few-shot) & 25 & 28 & 53 & 0.683 \\
		\midrule
		LaBSE-SVM    & LaBSE-LR & 3 & 5 & 8 & 0.508 \\
		\midrule
		TF-IDF-LR    & LaBSE-LR      & 19 & 32 & 51 & 0.0704 \\
		\midrule
		TF-IDF-SVM   & LaBSE-SVM     & 21 & 32 & 53 & 0.134 \\
		\bottomrule
	\end{tabular}
\end{table}

\subsection{Ablation Studies}
\label{sec:ablation-study}

We conduct two ablation studies to examine (1) the  impact of data augmentation across embedding families and architectures, and (2) the effect of fine-tuning pretrained embeddings on the electoral commitment dataset. For the study, we consider the improvement of $\geq$1.00\% in the $\Delta$ macro $F_1$-score meaningful as this percent value corresponds to approximately five correctly classified instances on our 486 instance test set. Improvements smaller than this threshold are considered likely to result from sampling noise rather than genuine model improvement.

\subsubsection{Impact of Data Augmentation}

\label{sec:ablation-augmentation}

Table~\ref{tab:augmentation_ablation} presents the performance of selected models trained with and without the 155 reviewed NC instances added to the training set. The results reveal a definite pattern across embedding families and architectures. TF-IDF-LR had the highest gains from augmentation across all metrics with macro $F_1$-score improving by +2.30\%. Among the fastText-based architectures, fastText-LSTM benefited the most (+2.40\% macro $F_1$-score) which is followed by fastText-GRU (+1.60\% macro $F_1$-score) and fastText-CNN (+1.10\% macro $F_1$-score). XLM-R also showed a modest improvement (below threshold) of +0.80\% macro $F_1$.

In contrast, the best performing LaBSE-based models showed little to no sensitivity to augmentation. LaBSE-LR showed no improvement whatsoever ($\Delta$ = 0.00\% across all metrics), and LaBSE-SVM gained only +0.60\% macro $F_1$ (below threshold).

Overall, these findings suggest that the benefit of data augmentation is closely tied to the embedding type rather than the architecture. Sparse embeddings benefit meaningfully from additional training samples, while large externally pretrained sentence encoders such as LaBSE show little sensitivity to class imbalance correction through augmentation. 

\begin{table}[H]
	\centering
	\caption{Ablation Study 1: Impact of data augmentation 
		across embedding families and architectures. 
		\iteng{w augmentation} denotes training with 155 
		synthetic-reviewed NC instances added to the training set, 
		\iteng{w/o augmentation} denotes training on original 
		data only. $\Delta$ indicates change over the 
		no-augmentation baseline. Meaningful 
		improvement are \bfeng{bolded ($\geq 1.00\%$)}.}
	\label{tab:augmentation_ablation}
	\begin{tabular}{l  c  c  c  c  c}
		\toprule
		Model & Accuracy & 
		ma-Precision  & 
		ma-Recall & 
		ma-$F_1$-score & 
		AUROC \\
		\midrule
		TF-IDF-LR \iteng{w augmentation}
		& 0.870 & 0.869 & 0.869 & 0.869 & 0.943 \\
		TF-IDF-LR \iteng{w/o augmentation} & 0.848 & 0.846 & 0.846 & 0.846 & 0.938 \\
		\midrule
		$\Delta$TF-IDF-LR & \bfeng{+2.20\%} & \bfeng{+2.30\%} & \bfeng{+2.30\%} & \bfeng{+2.30\%} & \bfeng{+0.50\%} \\
		\midrule
		LaBSE-LR \iteng{w augmentation}
		& 0.897 & 0.896 & 0.896 & 0.896 & 0.947 \\
		LaBSE-LR \iteng{w/o augmentation} & 0.897 & 0.896 & 0.896 & 0.896 & 0.947 \\
		\midrule
		$\Delta$LaBSE-LR & 0.00\% & 0.00\% & 0.00\% & 0.00\% & 0.00\% \\
		\midrule
		LaBSE-SVM \iteng{w augmentation}
		& 0.893 & 0.891 & 0.893 & 0.892 & 0.952  \\
		LaBSE-SVM \iteng{w/o augmentation} & 0.887 & 0.885 & 0.886 & 0.886 & 0.952 \\
		\midrule
		$\Delta$LaBSE-SVM & +0.60\% & +0.60\% & +0.70\% & +0.60\% & 0.00\% \\
		\midrule
		fastText-LSTM \iteng{w augmentation}
		& 0.914 & 0.912 & 0.915 & 0.913 & 0.961 \\
		fastText-LSTM \iteng{w/o augmentation} & 0.891 & 0.891 & 0.888 & 0.889 & 0.951 \\
		\midrule
		$\Delta$fastText-LSTM & \bfeng{+2.30\%} & \bfeng{+2.10\%} & \bfeng{+2.70\%} & \bfeng{+2.40\%} & \bfeng{+1.00\%} \\
		\midrule
		fastText-GRU \iteng{w augmentation}
		& 0.905 & 0.904 & 0.904  & 0.904 & 0.959 \\
		fastText-GRU \iteng{w/o augmentation} & 0.889 & 0.887 & 0.888 & 0.888 & 0.955 \\
		\midrule
		$\Delta$fastText-GRU & \bfeng{+1.60\%} & \bfeng{+1.70\%} & \bfeng{+1.60\%} & \bfeng{+1.60\%} & \bfeng{+0.40\%} \\
		\midrule
		fastText-CNN \iteng{w augmentation}
		& 0.889 & 0.887 & 0.890 & 0.888 & 0.959 \\
		fastText-CNN \iteng{w/o augmentation} & 0.879 & 0.877 & 0.877 & 0.877 & 0.958 \\
		\midrule
		$\Delta$fastText-CNN & \bfeng{+1.00\%} & \bfeng{+1.00\%} & \bfeng{+1.30\%} & \bfeng{+1.10\%} & \bfeng{+0.10\%} \\
		\midrule
		XLM-R \iteng{w augmentation}
		& 0.883 & 0.881 & 0.886 & 0.882 & 0.940 \\
		XLM-R \iteng{w/o augmentation} & 0.875 & 0.874 & 0.878 & 0.874 & 0.933 \\
		\midrule
		$\Delta$XLM-R & +0.80\% & +0.70\% & +0.80\% & +0.80\% & +0.70\% \\
		
		\bottomrule
	\end{tabular}
\end{table}

\subsubsection{Impact of Fine-tuning Pretrained Embeddings}

\label{sec:ablation-finetuning}

Table~\ref{tab:pretrain_ablation} compares the performance of pretrained embeddings (frozen form) against versions further fine-tuned on the training corpus. For fastText-based models, the fine-tuned version showed substantial improvements across all architectures. fastText-LSTM showed the highest gain with macro $F_1$-score improving by 6.40\% over the pretrained baseline which is followed by fastText-GRU (+5.30\% macro $F_1$-score) and fastText-CNN (+1.70\% macro $F_1$-score).

For LaBSE-based models, however, fine-tuning on the training corpus produced no meaningful improvement. LaBSE-LR showed a marginal drop of 0.20\% macro $F_1$ after fine-tuning and LaBSE-SVM showed no change at all ($\Delta$ = 0.00\% macro $F_1$-score).

Taken together, these findings highlight a clear distinction between lightweight and large pretrained embedding models in the context of low-resource domain adaptation. Lightweight models such as fastText benefit  from corpus-specific fine-tuning, while large pretrained encoders such as LaBSE are already better for this task without further adaptation.

\begin{table}[H]
	\centering
	\caption{Ablation Study 2: Impact of fine-tuning 
		pretrained embeddings on the electoral commitment 
		dataset. \iteng{fine-tuned} denotes pretrained 
		embeddings further trained on the electoral commitment 
		training corpus, \iteng{pre-trained} denotes frozen 
		pretrained embeddings used without any domain 
		adaptation. $\Delta$ indicates change of fine-tuned 
		over pre-trained baseline.Meaningful 
		improvement are \bfeng{bolded ($\geq 1.00\%$)}.}
	\label{tab:pretrain_ablation}
	\begin{tabular}{l  c  c  c  c  c}
		\toprule
		Model & Accuracy & 
		ma-Precision  & 
		ma-Recall & 
		ma-$F_1$-score & 
		AUROC \\
		\midrule
		fastText-LSTM \iteng{(fine-tuned)}
		& 0.914 & 0.912 & 0.915 & 0.913 & 0.961 \\
		fastText-LSTM \iteng{(pre-trained)} & 0.850 & 0.848 & 0.852 & 0.849 & 0.922 \\
		\midrule
		$\Delta$fastText-LSTM & \bfeng{+6.40\%} & \bfeng{+6.40\%} & \bfeng{+6.30\%} & \bfeng{+6.40\%} & \bfeng{+3.90\%} \\
		\midrule
		fastText-GRU \iteng{(fine-tuned)}
		& 0.905 & 0.904 & 0.904  & 0.904 & 0.959 \\
		fastText-GRU \iteng{(pre-trained)} & 0.852 & 0.850 & 0.854 & 0.851 & 0.934 \\
		\midrule
		$\Delta$fastText-GRU & \bfeng{+5.30\%} & \bfeng{+5.40\%} & \bfeng{+5.00\%} & \bfeng{+5.30\%} & \bfeng{+2.50\%} \\
		\midrule
		fastText-CNN \iteng{(fine-tuned)}
		& 0.889 & 0.887 & 0.890 & 0.888 & 0.959 \\
		fastText-CNN \iteng{(pre-trained)} & 0.872 & 0.870 & 0.873 & 0.871 & 0.942 \\
		\midrule
		$\Delta$fastText-CNN & \bfeng{+1.70\%} & \bfeng{+1.70\%} & \bfeng{+1.70\%} & \bfeng{+1.70\%} & \bfeng{+1.70\%} \\
		\midrule
		LaBSE-LR \iteng{(fine-tuned)} & 0.895 & 0.893 & 0.894 & 0.894 & 0.950 \\
		LaBSE-LR \iteng{(pre-trained)}
		& 0.897 & 0.896 & 0.896 & 0.896 & 0.947 \\
		\midrule
		$\Delta$LaBSE-LR & -0.20\% & -0.30\% & -0.20\% & -0.20\% & +0.30\% \\
		\midrule
		LaBSE-SVM \iteng{(fine-tuned)} & 0.893 & 0.891 & 0.893 & 0.892 & 0.948 \\
		LaBSE-SVM \iteng{(pre-trained)}
		& 0.893 & 0.891 & 0.893 & 0.892 & 0.952 \\
		\midrule
		$\Delta$LaBSE-SVM & 0.00\% & 0.00\% & 0.00\% & 0.00\% & -0.40\% \\
		
		\bottomrule
	\end{tabular}
\end{table}

\subsection{Error Analysis}
\label{sec:error-analysis}

\begin{figure}[h]
	\centering
	\includegraphics[width=0.40\textwidth]{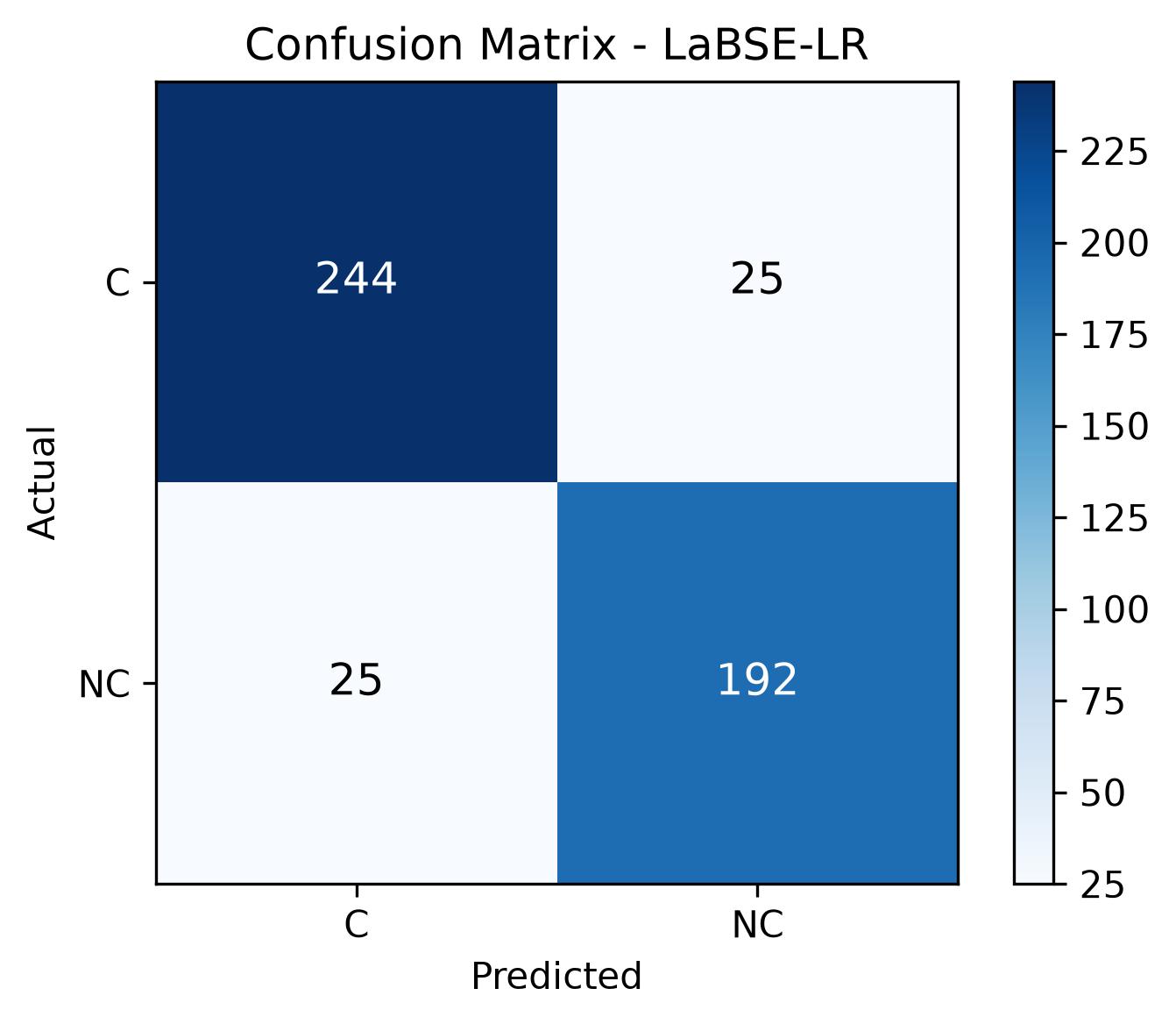}
	\includegraphics[width=0.40\textwidth]{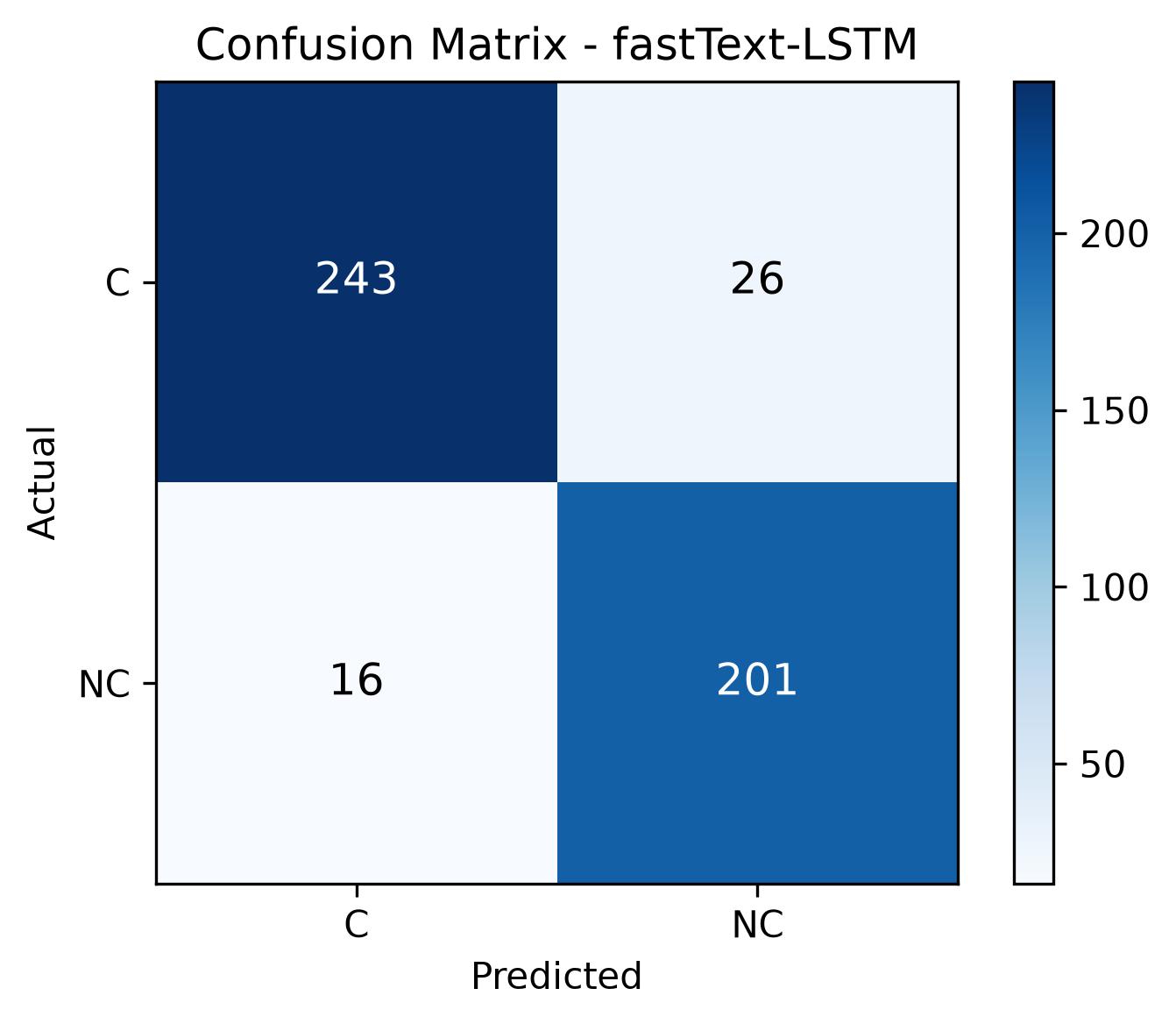}
	\includegraphics[width=0.40\textwidth]{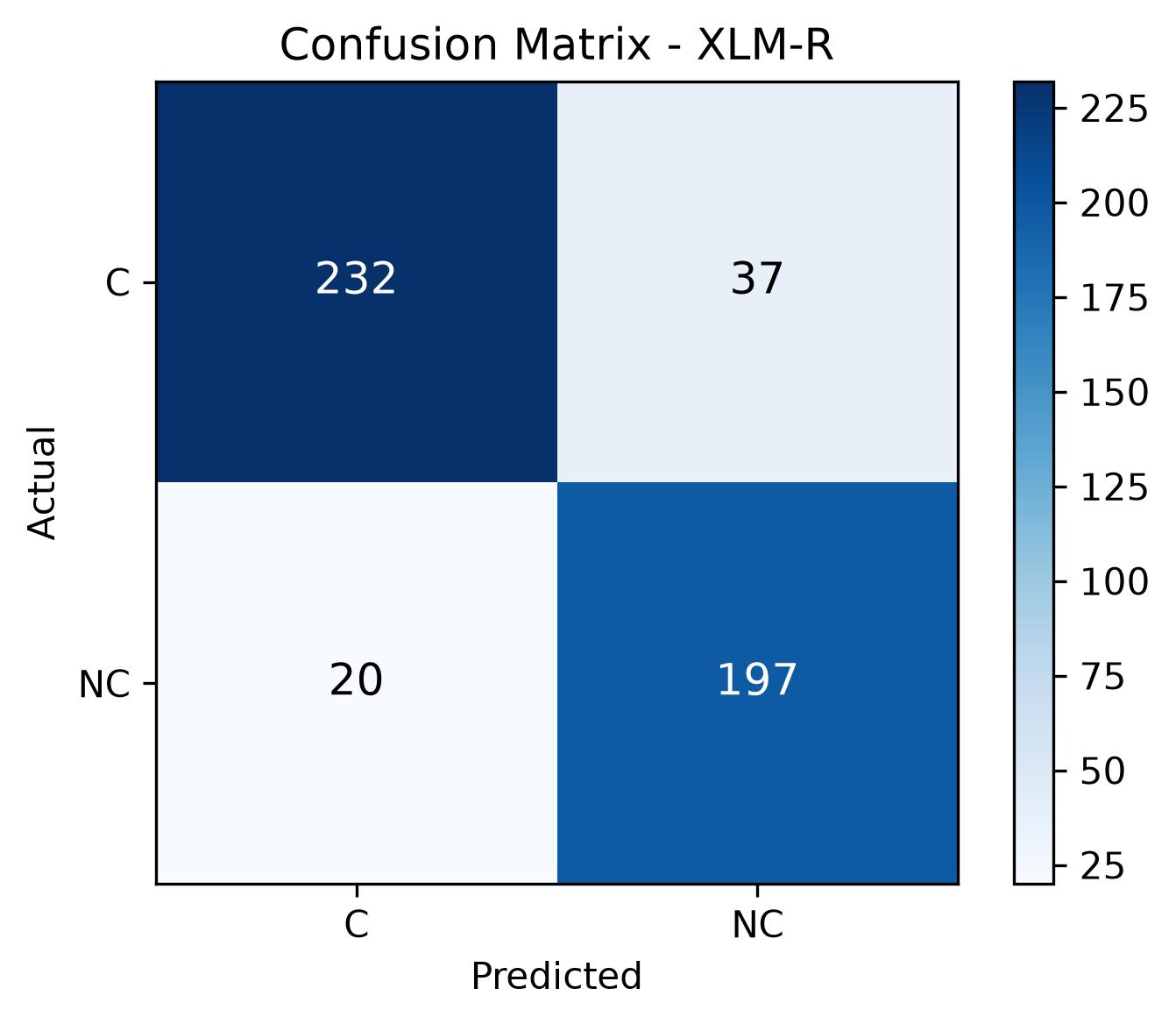}
	\includegraphics[width=0.40\textwidth]{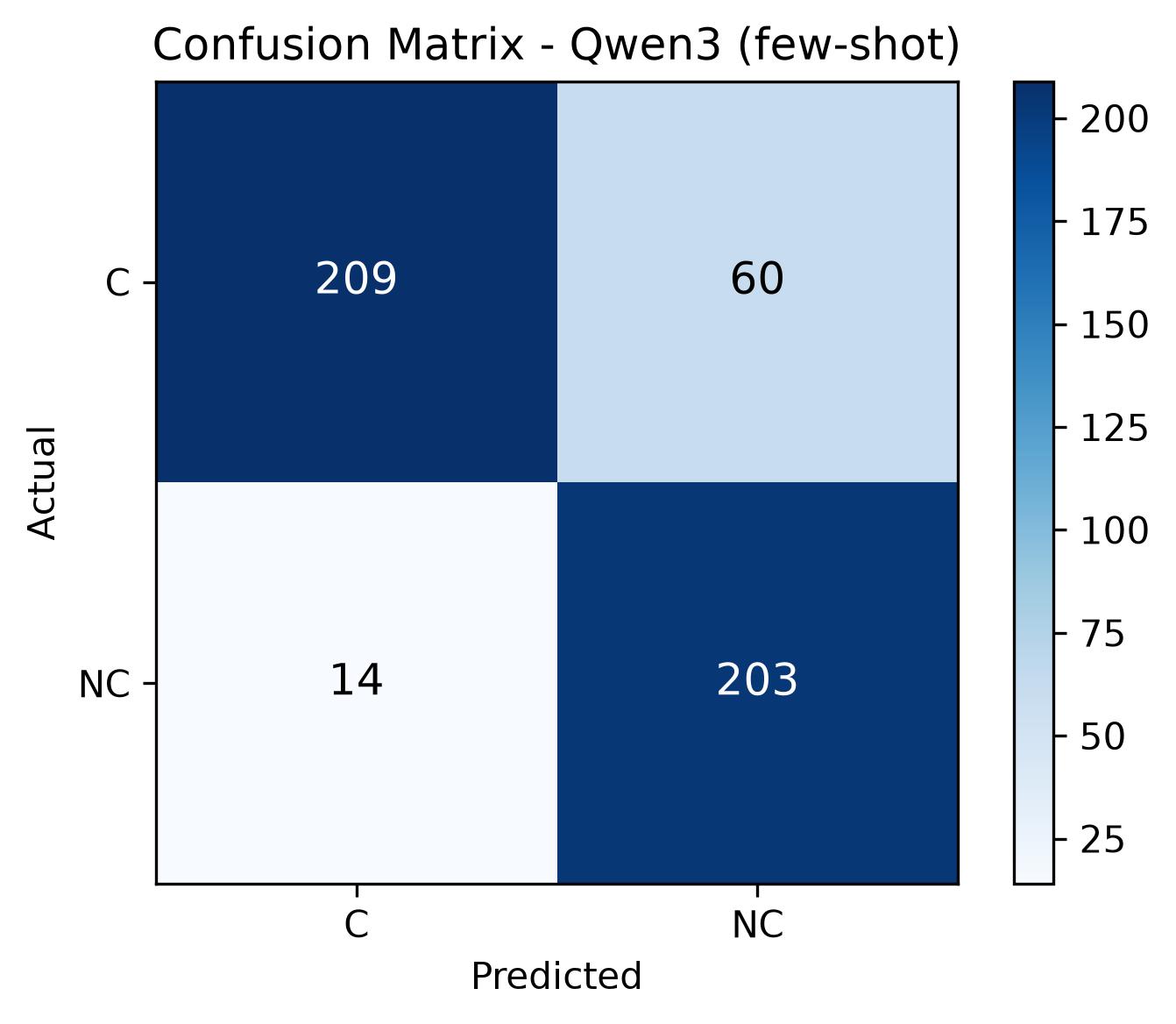}
	\caption{Confusion matrices for LaBSE-LR, 
		fastText-LSTM, XLM-RoBERTa, and Qwen3-8B (few-shot).}
	\label{fig:confusion_matrices}
\end{figure}

We performed a detailed qualitative analysis of the errors made by the four strongest models from each methodological family: LaBSE-LR, fastText-LSTM,  XLM-R and Qwen3-8B (few-shot). The analysis draws on the statements that the models misclassified. 
Figure~\ref{fig:confusion_matrices} presents the 
confusion matrices for the four models.

There are two broad tendencies that are immediately visible across the confusion matrices. First, false negatives outnumber false positives in every model and second, the imbalance grows steadily as we move from classical embedding-based classifiers to larger multilingual models. Qwen3-8B is the most conservative system as it produces the fewest false positives, but it has more than twice as many false negatives as LaBSE-LR.

As observed, the errors across all four models cluster around three interrelated linguistic factors. They are listed below:

\begin{itemize}[label={}]
	\item {\bfeng{Lexical cue:}} The first factor is that the models reliance on specific words such 
	as \iteng{\nep{"प्रोत्साहन"}} (encouragement), \iteng{\nep{"पहल"}} (initiative), 
	\iteng{\nep{"अवलम्बन"}} (adopt), \iteng{\nep{"सहजिकरण"}} (to facilitate), and 
	\iteng{\nep{"परिचालन"}} (mobilize). The models treat these words as sufficient evidence to predict the statement as a commitment, instead of actually checking whether the predicate they form specifies anything concrete, measurable, and deliverable. So, in cases where statements are vague and aspirational (non-commitements) such cues trigger them into false positive results.

	{\bfeng{Example:}} \nep{युवाहरुलाई सीपमूलक तालिमको व्यवस्था गरी लघु तथा घरेलु उद्योग सञ्चालनमा प्रोत्साहन गरिने छ।} {\iteng{`Youth will be encouraged to operate small and cottage industries by providing skill-based training.'}}
	
	The above statement has a canonical future-tense form and the phrase \nep{"सीपमूलक तालिमको व्यवस्था गरी"} (provide skill-based training) looks concrete but the main predicate remains just pure encouragement with no mention of either scale, budget or delivery mechanism. Hence, the true label of this statement is NC. The model seems to combine the future-tense marker with this surface-level concreteness and miss that the actual predicate is weak. This pattern shows up in all six test cases, where the word \nep{"प्रोत्साहन"} (encouragement) is what appears to influence the prediction toward C.
	
	\item {\bfeng{Syntactic completeness:}} The second observed factor is that telegraphic and fragment-like statements are assigned frequently to the NC class even when their content is fully specific and their true label is C. These statements often end in a non-finite or non-canonical verb form instead of the finite future-tense marker (\nep{"गर्ने छ"}, \nep{"गरिने छ"}) that models predict correctly.
	
	{\bfeng{Example:}} \nep{ईन्टरनेट सेवा जिल्ला भर} {\iteng{`Internet service throughout the district'}}
	
	The above statements promises a district-wide internet provision which is concrete commitment as it specifies a clear scope and target beneficiary population but lacks a canonical future tense form which influence the models to predict NC. Such factor have been seen in 24 cases where at least two model have misclassified the statements.
	
	\item {\bfeng{Information density:}} The third factor that we observed is the information density and length of the statements.  In our case, the extremely long statements ($\geq 30$) have diluted the main predicate signals, resulting in misclassifications.
	
	{\bfeng{Example:}} \nep{चुनौती तथा समस्यालाई सम्भावनाहरूमा बदल्न जिम्मे वार पार्टी हुनुको नाताले नेकपा एमाले आफ्ना पार्टीका प्रतिनिधिहरू र स्थानीय सरकारसँग सम्बन्ध कायम गरी नगरपालिका विकासमा समन्वय, सहजीकरण र नियन्त्रण तथा निर्देशनसमेत गर्न पर्सागढी नगरपालिका कमिटी निम्नानुसार प्रतिबद्धता व्यक्त गर्दछौं}। {textit{`As a party responsible for turning challenges and problems into opportunities, the CPN-UML, by maintaining relations with its party representatives and local governments, we, the Parsagarhi Municipality Committee, express our commitment as follows to coordinate, facilitate, control and direct the development of the municipality.'}}
	
	The above statement is 37-word and have long stacks of several clauses from turning challenges into opportunities, maintaining relations with party representatives and local government to coordinating and facilitating municipal development. With such intervening content, the main predicate signal gets diluted and the model appears to lose track of it by the time it reaches the end of the sentence which leads to misclassification. There are a total of six cases where such long sentences are classified incorrectly.

\end{itemize}

\begin{table*}[h!]
	\centering
	\caption{Representative example error cases grouped by 
		linguistic factor. FP = false positive (NC predicted 
		as C), FN = false negative (C predicted as NC).}
	\label{tab:error_examples}
	\begin{tabular}{p{3cm} p{4cm} p{3.5cm} c}
		\toprule
		Factor & 
		Statement (ne) & 
		Statement (en) &
		Error Type \\
		\midrule

		& \nep{युवाहरुलाई सीपमूलक तालिमको व्यवस्था 
			गरी लघु तथा घरेलु उद्योग सञ्चालनमा प्रोत्साहन 
			गरिने छ।} 
		&	Youth will be encouraged to operate small and cottage industries by providing skill-based training.
		& FP \\
		\cmidrule{2-4}
		
		Lexical cue & \nep{प्रत्येक तहमा जिम्मेवारी विने, कामको 
			मुल्याङ्‌कन गर्ने र पदोन्नति गर्ने वैज्ञानिक 
			पद्धतिको निर्माण गरी अवलम्वन गरिने छ।}
			&  A scientific system for assigning responsibilities, evaluating work, and promoting will be developed and adopted at each level.
		& FP \\
		\cmidrule{2-4}
		
		& \nep{नगरभित्र कवड हल र रतुवामाई ४ गढीमाई 
			महाराजथानमा रंगशाला निर्माणका लागि आवश्यक 
			पहल गरिने छ।} 
			& Necessary initiatives will be taken to construct a stadium in the city's Kawad Hall and Ratuwamai 4 Gadhimai Maharajthan.
		& FP \\

		\midrule

		& \nep{ईन्टरनेट सेवा जिल्ला भर} 
		& Internet service throughout the district
		& FN \\
		\cmidrule{2-4}
		
		& \nep{उदाहरणीय सुशासन} 
		& Exemplary Good Governance
		& FN \\
		\cmidrule{2-4}
		
		Syntactic completeness	& \nep{जस अन्तरगत ५ वर्ष भित्र १ लाख नयाँ 
			बोट रोपनि छन} 
			& Under which 1 lakh new trees will be planted within 5 years.
		& FN \\
		\cmidrule{2-4}
		
		& \nep{समग्र मानव विकासका लागि ठोस कार्ययोजना तय गरी कार्यान्वयन गर्ने} 
		& To formulate and implement a concrete action plan for overall human development
		& FN \\
		\cmidrule{2-4}
		
		& \nep{प्रेरित गर्न बजारको माग अनुसारको प्राविधिक सौप}
		& Technical handover according to market demand to motivate 
		& FN \\
		
		\midrule

		& \nep{चुनौती तथा समस्यालाई सम्भावनाहरूमा बदल्न जिम्मे वार पार्टी हुनुको नाताले नेकपा एमाले आफ्ना पार्टीका प्रतिनिधिहरू र स्थानीय सरकारसँग सम्बन्ध कायम गरी नगरपालिका विकासमा समन्वय, सहजीकरण र नियन्त्रण तथा निर्देशनसमेत गर्न पर्सागढी नगरपालिका कमिटी निम्नानुसार प्रतिबद्धता व्यक्त गर्दछौं} 
		& As a party responsible for turning challenges and problems into opportunities, the CPN-UML, by maintaining relations with its party representatives and local governments, we, the Parsagarhi Municipality Committee, express our commitment as follows to coordinate, facilitate, control and direct the development of the municipality.
		& FN \\
		\cmidrule{2-4}

		Information density & \nep{सिचाइ सुविधा पुगेका वडाहरुका सबै 
			किसानका खेतवारीमा सिंचाइको व्यवस्था गर्न 
			कुलो नहर लाई व्यवस्थित गर्दै र सिचाई सुविधा 
			नपुगेका वडाहरुमा तत्काल वैकल्पिक र दिर्घकालिन 
			सिंचाइको व्यवस्थाका लागि पहल गरी बाहै महिना 
			सिंचाई सुविधा उपलव्ध गराई बाटै महिना खेती 
			गर्न प्रोत्साहन गरिनेछ।} 
			& To provide irrigation facilities to all farmers in wards with irrigation facilities, canals will be organized and initiatives will be taken to provide alternative and long-term irrigation facilities in wards without irrigation facilities, and monthly cultivation will be encouraged by providing irrigation facilities for half a year.
		& FP \\
		
		\bottomrule
	\end{tabular}
\end{table*}

\section{Discussion}
\label{sec:discussion} 

The results indicate that DL models achieve higher performance than classical ML models in our binary classification of political texts in Nepali. Among all evaluated configurations, fastText-LSTM achieved the best overall performance of macro $F_1$-score = 0.913, while LaBSE-LR emerged as the best ML model with macro $F_1$-score = 0.896. We applied McNemar’ test between these models to compare the difference in their performance. Even when the discordant pairs where LSTM was correct and LR was wrong was greater than the pairs where LR was correct and LSTM was wrong, the result, however, showed no statistically significant difference between them, suggesting that the two approaches achieve equivalent performance on this task.  This is a notable finding from a deployment perspective: fastText-LSTM contains only $\approx$ 2.9M parameters and requires 19.54 MB of GPU memory to load for inference, compared to the $\approx$ 471M parameter LaBSE encoder \citep{feng2022language} which requires 1,798.98 MB to load . For resource-constrained deployment settings such as mobile or other edge applications, fastText-LSTM is therefore the more practical choice.

Among the ML models, LaBSE embeddings showed higher performance than TF-IDF across all model families, although the difference did not reach statistical significance for best performing models, LR and SVM, both falling short by narrowest of margins. This near-equivalence is consistent with prior findings that TF-IDF representations remain competitive with dense contextual embeddings on smaller or more lexically-driven datasets, since sparse lexical features can effectively capture task-relevant surface patterns without requiring the larger training data typically needed to fully exploit dense embedding spaces \citep{sazan2024enhancing, abdelmotaleb2025word}. This also suggests that for Nepali commitment mining, explicit lexical cues carry much of the discriminative signal, narrowing the advantage that language-agnostic sentence embeddings like LaBSE \citep{feng2022language} might otherwise provide on more semantically complex tasks. 

A key finding from the DL experiments is that fastText-LSTM showed a statistically significant advantage over both transformer-based models (see Table~\ref{tab:dl_model_performance}). This aligns with the observation of \citet{ezencan2020comparisonlstmbertsmall} that simple bidirectional LSTM models can achieve significantly higher results than BERT-based models on small datasets. Our dataset of 2,585 instances (with augmentation) is within the range where transformer models are known to underperform simpler recurrent architectures due to their high parameter counts and dependence on large pretraining corpora for effective fine-tuning. The morphological complexity of Nepali further contributes to this gap. \citet{jha2024exploring} highlights that Nepali's complex morphology and script pose unique tokenization challenges, and the WordPiece tokenization used by mBERT and XLM-R is not specifically optimized for Devanagari script morphology. In contrast, fastText's character n-gram approach naturally captures subword morphological patterns, which is advantageous for morphologically rich languages \citep{bojanowski2017enriching}. Among the transformer models, XLM-R showed slightly higher performance than mBERT, which is consistent with the findings of \citet{conneau2019unsupervised}. However, McNemar's test showed no statistically significant difference between them, suggesting that for our specific task and dataset size, neither transformer architecture offers a meaningful advantage over the other.

The ablation studies revealed further insights into the role of data augmentation and embedding adaptation for this task. In the first study, we observed that the benefit of augmentation is closely linked to the embedding type where sparse TF-IDF representations and domain fine-tuned fastText embeddings benefit from additional training samples, while the rich multilingual representations of LaBSE provide sufficient generalization for the mild class imbalance present in our dataset and does not benefit much from additional augmentation (see Table~\ref{tab:augmentation_ablation}). In the second study, we observed that domain fine-tuning of fastText embeddings on the commitment mining dataset (train split) showed substantial improvements across all fastText-based architectures (see Table~\ref{tab:pretrain_ablation}). This indicates that pretrained models trained on Nepali web text from Common Crawl do not adequately represent the distinct lexical characteristics, specific future-tense constructions, formal bureaucratic phrasing, and domain-specific nouns found in electoral manifestos and political pamphlets, and that domain adaptation is essential for optimal performance. In contrast, fine-tuning LaBSE on the same corpus produced no meaningful improvement, which is expected given that fine-tuning a 471M parameter model on fewer than a small set of training sentences is unlikely to produce reliable domain adaptation. 

The comparison of lightweight fine-tuned models with large language models reveals an important finding in our task. Despite the remarkable general-purpose capabilities and larger parameter size of Qwen3-8B and Llama-3.1-8B-Instruct, they both fell to perform to the level of lightweight models. This supports the work of \citet{bucher2024fine}, who showed that fine-tuning smaller models still significantly outperforms zero-shot prompted generative AI models on text classification tasks. Few-shot prompting showed a statistically significant improvement over zero-shot for Qwen3-8B, confirming that in-context examples help guide LLMs toward the task-specific distinction between commitment and non-commitment language. However, the performance gap relative to fine-tuned models suggests that even a few examples are insufficient to fully capture broad spectrum of lexical characteristics that  domain-specific fine-tuning on labelled dataset can, which is reliable commitment classification in Nepali political text. This directly justifies the dataset creation effort of this work. 

The error analysis further reveals why certain error patterns persist across all model families. The three dominant factors, lexical cues, syntactic completeness, and information density, indicate that all evaluated models, from LaBSE-LR to Qwen3-8B, rely on surface morphological and lexical cues rather than on a deeper representation of specificity and actionability. The shared false positives driven by lexical cues suggest that models treat future-tense morphology and domain nouns as sufficient evidence of commitment, while remaining insensitive to the absence of concrete mechanisms or measurable outcomes. This connects directly to the annotation challenges discussed in Section~\ref{sec:data-annotation}, where our expert annotator reclassified several vague statements as non-commitments precisely because they lacked specific actionable content. The fact that models fail on the same cases that challenged human annotators suggests these are genuinely hard instances that may require richer linguistic representations beyond what current models offer. Addressing this gap may require either augmenting the training data with more diverse manifesto styles or developing models that explicitly capture lexical  structures, forms, and vocabulary characteristic of telegraphic and organizational styles found in real Nepali political manifesto text.

\section{Limitations} 
\label{sec:limitations}

Although this research presents a meaningful step forward in Nepali commitment mining through its dataset, experiments, and application, we would like to acknowledge several limitations that should considered when interpreting the results.

\begin{itemize}
	\item We trained the classification models on a  small dataset of 2,585 instances, which may have limited performance, especially for the deep learning and transformer-based models that usually need more data to perform well.
	
	\item We evaluated the generated images internally, with no outside expert evaluation. Given the small subset, a systematic ablation on prompt engineering was outside the scope of this study. We instead performed an informal comparison of prompt variants.
	
	\item We did not establish a human performance 
	baseline for the commitment classification task. 
	While the inter-annotator agreement of $\kappa$ 
	= 0.777 provides an indirect measure of task 
	difficulty, a formal human baseline, where 
	annotators classify held-out test instances 
	independently, would provide a more direct 
	upper bound for model performance.
	
	\item We consider the mobile app at this stage to be just a working prototype and it has not gone through any formal usability testing among voters/users. 
	\item For the ablation and error analysis, we represented the major embedding families and architectures as much as possible, but we did not perform the analysis for all models evaluated in the study so, their behaviour remains unexplored.
	
	\item We collected most of the data through online sources and manual field visits, so commitments made informally or at the grassroots level might not be fully represented. We also collected audio and video data but did not process it as it was outside the current scope of this work. 
\end{itemize}

\section{Ethical Considerations}
\label{sec:ethics}

Since this work deals with electoral commitment text and involves human judgment on potentially sensitive content, we briefly discuss some ethical concerns related to annotator bias, data sources, and the possible impact of the system. Our annotators come from media and public communication backgrounds, which helped them understand political language better, but this also means they may carry unconscious biases when labeling borderline statements. We reduced the potential influence of political affiliations by having two independent annotators and a blind adjudication step (see Section~\ref{sec:data-annotation}), where annotators and the adjudicator were provided only with the statements and did not have access to metadata such as candidate names, political parties, or regions. This helped to make the labels determined by the content of the statements rather than the political identity of the candidates. However, considering that the annotation process was still based on human judgment, we cannot fully rule out other sources of subjective bias, like differences of interpretation, previous beliefs, or perspectives. Also, the dataset itself is not balanced across parties or provinces, as its composition depended on the availability of candidates’ publicly accessible materials (see Section~\ref{sec:data-collection}). Therefore, the results should not be interpreted as a direct comparison or evaluation of one party or candidate against another.

We also acknowledge that our data collection did not explicitly account for the terms and conditions of the platforms from which the data were collected. All data were collected manually from publicly accessible profiles for research purposes only. We did not misuse any candidate’s personal profile information or name beyond what they had already made public as part of their political campaign. This information was used only to correctly identify and attribute commitments to the appropriate candidate, and not for any other purpose. The final dataset contains only processed textual data and does not include any images or videos from the original platforms. Still, we acknowledge that platform policies may restrict the collection and use of publicly accessible content, including for research purposes, and future large scale collection should check this more carefully or get proper permission.

In terms of impact, the system is meant to help citizens track election 
promises, but there are some risks too. The GAI-generated images 
(see Section~\ref{sec:GAI}) are only illustrative and not real depictions 
of what will actually happen, and if a user is not careful this could 
be misread as evidence something has already been done. Similarly since our classifier is not perfect (see Section~\ref{sec:experiments_results}), 
some real commitments may get missed (false negative) or some vague 
statement may get shown as commitment (false positive), both are 
problematic in a system whose whole point is accountability, so results 
from this prototype should not be taken as final/verified judgement on 
any candidate. Lastly, since only a subset of candidates/provinces are 
currently covered, some candidates get visibility in the app while 
others don't, simply due to data availability and not their actual 
performance, which is something we aim to address through wider data 
collection in future work (see Section~\ref{sec:conclusion}).

\section{Conclusion and Future Work}
\label{sec:conclusion}
Our research contributes to strengthen democratic accountability by developing a system that systematically identifies, structures, and analyzes electoral commitments made by candidates. Since political promises articulated in manifestos, campaign speeches, and other publicity materials often fade into obscurity once elections conclude which undermine citizen trust in political processes, this research addresses that gap through the development of a deep learning  model supported by a gold-standard dataset of electoral commitments. By performing comparative analysis between both ML, DL, and LLM methods, the study identifies effective approaches for commitment classification in the Nepali political context. The result of the comparative analysis shows fastText-LSTM as the best model both in terms of performance and deployment aspects.

We also integrate the model into a mobile application to demonstrate the practical implications of our research. Through an accessible, user-friendly interface, the civils can obtain concise summaries of candidate commitments, supported by GAI images for enhanced interpretability. Such a system provides citizens with a tangible tool to evaluate whether promises are honored which promotes transparency and encouraging elected officials to remain accountable to their constituencies. Beyond its technical contributions, our research advances the broader democratic goal of promoting political trust, transparency, and civic engagement.

Despite these contributions, our research has certain limitations that we discussed in Section~\ref{sec:limitations}. In future work, we plan to work on these limitations from collecting contextually rich commitment data from diverse multimedia sources to developing, deploying and testing a robust application among voters and users. Additionally, we plan to prioritize extensive field data collection during election periods in various regions of Nepal to capture a wider spectrum of political discourse. We plan to conduct field visits and systematically document commitments from all candidates at the local level which would enhance the coverage and reliability of the dataset discussed in Section \ref{sec:methods and materials}. This advancement would enable citizens to access localized candidate information, track electoral promises in real time, and ultimately strengthen democratic participation and accountability in Nepal.

\paragraph{Acknowledgments}
We sincerely thank our student team members Mr. Mandip Kumar Thakur and Mr. Aayush Shrestha  for their valuable contributions, and we are grateful to our expert advisors for their insightful guidance. We also acknowledge the efforts of our data annotators and the collaborative support of political stakeholders in facilitating data collection and annotation. Finally, we extend our appreciation to the ILPRL lab team for their assistance and support throughout this research.

\paragraph{Funding Statement}
This research was funded by Kathmandu University Research, Development, and Innovation (KU RDI).

\paragraph{Competing Interests}
None

\paragraph{Data \& Supplementary Materials Availability Statement}
The dataset, source code, annotation guidelines, and annotator review sheets are publicly available at \\
\verb+\url{https://github.com/Nischal-Karki-1/from-promises-to-accountability}+. 

\paragraph{Ethical Standards}
This research adheres to all applicable ethical guidelines and legal requirements in Nepal. Data collection was conducted transparently with publicly available political materials, ensuring no personal or sensitive information was misused. All annotators participated following a standardized guidelines, and the study maintained integrity and accountability throughout the research process.
\paragraph{Author Contributions}
Conceptualization: P.P. 
Data curation: P.P.; N.K.; S.T.; R.C.; A.D. 
Formal analysis: P.P.; N.K. 
Funding acquisition: P.P. 
Investigation: N.K. 
Methodology: P.P.; N.K.; B.K.B. 
Project administration: P.P. 
Resources: P.P. 
Software: N.K. 
Supervision: P.P. 
Validation: P.P.; B.K.B. 
Visualization: P.P.; N.K. 
Writing – original draft: N.K.; P.P. 
Writing – review \& editing: P.P.; N.K.; A.D.; R.C.; S.T.; B.K.B. 
All authors approved the final submitted version.

\defbibnote{preamble}

\printbibliography[prenote={preamble}]

\appendix

\section{Model Configuration and Hyperparameter Details}
\label{sec:appendix_hyperparams}

This appendix presents the optimal hyperparameter configurations identified for each model used in this study, reported in Tables~\ref{tab:ml_models_parameters} and \ref{tab:dl_models_parameters}.

\begin{table}[H]
	\centering
	\caption{Optimal hyperparameter settings for ML models under TF-IDF and LaBSE text representations}
	\label{tab:ml_models_parameters}
	\begin{tabular}{l c p{8cm}}
		\toprule
		Model & Text Representation & Hyperparameters \\
		\midrule
		
		\multirow{2}{*}{LR}
		& TF-IDF
		& C=10.0, class\_weight=None, penalty=l2, max\_features=20000, ngram\_range=(1, 2) \\
		& LaBSE
		& C=1.0, class\_weight=balanced, penalty=l2 \\
		
		\midrule
		
		\multirow{2}{*}{DT}
		& TF-IDF
		& class\_weight=None, criterion=entropy, max\_features=5000, ngram\_range=(1, 1), max\_depth=20, min\_samples\_split=10, min\_samples\_leaf=2 \\
		& LaBSE
		& class\_weight=balanced, criterion=entropy, max\_depth=5, min\_samples\_split=2, min\_samples\_leaf=1 \\
		
		\midrule
		
		\multirow{2}{*}{RF}
		& TF-IDF
		& class\_weight=None, n\_estimators=300, max\_features=5000, max\_depth=None, min\_samples\_split=2, min\_samples\_leaf=1, ngram\_range=(1, 2) \\
		& LaBSE
		& class\_weight=None, n\_estimators=100, max\_depth=10, min\_samples\_split=10, min\_samples\_leaf=2 \\
		
		\midrule
		
		\multirow{2}{*}{SVM}
		& TF-IDF
		& C=1.0, class\_weight=None, gamma=scale, max\_features=20000, kernel=linear, ngram\_range=(1, 2) \\
		& LaBSE
		& C=1.0, class\_weight=None, gamma=scale, kernel=rbf \\
		
		\midrule
		
		\multirow{1.5}{*}{NB}
		& TF-IDF
		& max\_features=20000, ngram\_range=(1, 2), alpha=0.5, norm=True \\
		& LaBSE
		& var\_smoothing=1e-09 \\
		
		\midrule
		
		\multirow{2}{*}{KNN}
		& TF-IDF
		& metric=cosine, n\_neighbors=11, weights=distance, max\_features=5000, ngram\_range=(1, 2) \\
		& LaBSE
		& metric=cosine, n\_neighbors=7, weights=distance \\

		\bottomrule
	\end{tabular}
	
	\iteng{Note:} All other parameters used default configuration for all models unless otherwise specified.
\end{table}

\begin{table}[H] 
	\centering
	\caption{Architecture and optimal hyperparameter settings for DL models under fastText, GloVe, and transformer-based representations}
	\label{tab:dl_models_parameters}
	\begin{tabular}{l c p{4cm} p{4cm}}
		\toprule
		Model & Text Representation & Model Architecture & Hyperparameters \\
		\midrule
		
		\multirow{2}{*}{SRN}
		& fastText 
		& Embedding (max\_length=64, dim=300), SRN (36 units), Sigmoid (1) 
		& Batch size=16, epochs=30, patience=3, lr=3e-4, dropout=0.3, l2\_reg=0.0005 \\
		& GloVe 
		& Embedding (max\_length=64, dim=300), SRN (48 units), Dense (32 units), Sigmoid (1) 
		& Batch size=16, epochs=30, patience=3, lr=3e-4, dropout=0.5, l2\_reg=0.001 \\
		\midrule
		
		\multirow{2}{*}{LSTM}
		& fastText 
		& Embedding (max\_length=64, dim=300), LSTM (36 units), Sigmoid (1) 
		& Batch size=16, epochs=30, patience=3, lr=3e-4, dropout=0.3, l2\_reg=0.0005 \\
		& GloVe 
		& Embedding (max\_length=64, dim=300), LSTM (48 units), Dense (24 units), Sigmoid (1) 
		& Batch size=16, epochs=30, patience=3, lr=5e-4, dropout=0.3, l2\_reg=0.0005 \\
		\midrule
		
		\multirow{2}{*}{GRU}
		& fastText 
		& Embedding (max\_length=64, dim=300), GRU (36 units), Sigmoid (1) 
		& Batch size=16, epochs=30, patience=3, lr=3e-4, dropout=0.3, l2\_reg=0.0005 \\
		& GloVe 
		& Embedding (max\_length=64, dim=300), GRU (36 units), Dense (24 units), Sigmoid (1) 
		& Batch size=16, epochs=30, patience=3, lr=5e-4, dropout=0.3, l2\_reg=0.001 \\
		\midrule
		
		\multirow{2}{*}{CNN}
		& fastText 
		& Embedding (max\_length=64, dim=300), Conv1D (96 filters, kernel=3), GlobalMaxPooling1D, Dense (128), Sigmoid (1) 
		& Batch size=32, epochs=30, patience=5, lr=5e-4, dropout=0.2, l2\_reg=0.0005 \\
		& GloVe 
		& Embedding (max\_length=64, dim=300), Conv1D (32 filters, kernel=3), GlobalMaxPooling1D, Dense (32), Sigmoid (1) 
		& Batch size=16, epochs=30, patience=5, lr=5e-4, dropout=0.3, l2\_reg=0.01 \\
		\midrule
		
		mBERT
		& WordPiece
		& 12 Transformer layers (768 hidden units), Sigmoid (1) 
		& Batch size=16, epochs=10, patience=3, lr=5e-5, weight\_decay=0.0 \\
		\midrule
		
		XLM-R
		& SentencePiece
		& 12 Transformer layers (768 hidden units), Sigmoid (1) 
		& Batch size=8, epochs=10, patience=3, lr=5e-5, weight\_decay=0.0 \\
		
		\bottomrule
	\end{tabular}
	
\iteng{Note:} All DL models used Adam optimizer, ReLU activation, and L2 regularization. Pretrained embeddings/weights (fastText, GloVe, mBERT, XLM-R) were fine-tuned end-to-end on the training split for all models. For TF-IDF vectorization and fastText/GloVe training, text was tokenized via whitespace splitting on the cleaned Devanagari text.
\end{table}

\end{document}